\documentclass[preprint,12pt]{elsarticle}

\usepackage{amssymb}
\usepackage{amsmath}

\usepackage{booktabs}
\usepackage[table]{xcolor}
\definecolor{Gray}{gray}{0.9}
\usepackage{caption} 
\usepackage{subcaption}
\usepackage{multirow}
\usepackage{cleveref} 
\usepackage{tikz} 
\usepackage{amsmath}
\usetikzlibrary{positioning, arrows.meta, shapes.geometric, fit, calc, backgrounds}
\usepackage{url}

\journal{Science of Remote Sensing}

\begin{document}

\begin{frontmatter}



\title{Near–Real-Time Conflict-Related Fire Detection in Sudan Using Unsupervised Deep Learning} 


\author[1]{Kuldip Singh Atwal\corref{cor1}}
\ead{katwal@gmu.edu}

\author[1]{Dieter Pfoser}
\ead{dpfoser@gmu.edu}

\author[2]{Daniel Rothbart}
\ead{drothbar@gmu.edu}

\cortext[cor1]{Corresponding author}

\affiliation[1]{organization={Geography and Geoinformation Science, George Mason University},
            addressline={4400 University Drive}, 
            city={Fairfax},
            postcode={22030}, 
            state={VA},
            country={United States}}

\affiliation[2]{organization={The Jimmy and Rosalynn Carter School for Peace and Conflict Resolution, George Mason University},
            addressline={4400 University Drive}, 
            city={Fairfax},
            postcode={22030}, 
            state={VA},
            country={United States}}            

\begin{abstract}

Ongoing armed conflict in Sudan highlights the need for rapid monitoring of conflict-related fire-affected areas. Recent advances in deep learning and high-frequency satellite imagery enable near–real-time assessment of active fires and burn scars in war zones. This study presents a near–real-time monitoring approach using a lightweight Variational Auto-Encoder (VAE)–based model integrated with 4-band Planet Labs imagery at 3 m spatial resolution. We demonstrate that these impacted regions can be detected within approximately 24 to 30 hours under favorable observational conditions using accessible, commercially available satellite data. To achieve this, we adapt a VAE-based model, originally designed for 10-band imagery, to operate effectively on high-resolution 4-band inputs. The model is trained in an unsupervised manner to learn compact latent representations of nominal land-surface conditions and identify burn signatures by quantifying changes between temporally paired latent embeddings. Performance is evaluated across five case studies in Sudan and compared against cosine distance, CVA, and IR-MAD using precision, recall, F1-score, and the area under the precision–recall curve (AUPRC) computed between temporally paired image tiles. Results show that the proposed approach consistently outperforms the other methods, achieving higher recall and F1-scores while maintaining viable precision in highly imbalanced fire-detection scenarios. Experiments with 8-band imagery and temporal image sequences yield only marginal performance gains over single 4-band inputs, underscoring the effectiveness of the proposed lightweight approach for scalable, near–real-time conflict monitoring.

\end{abstract}


\end{frontmatter}



\section{Introduction}
\label{sec1}

An armed conflict in Sudan, which began in April 2023, has resulted in widespread civilian harm, large-scale displacement, and severe destruction of infrastructure. Fighting between the Sudanese Armed Forces (SAF) and the Rapid Support Forces (RSF) has resulted in the deaths of thousands of people and the displacement of over 12 million individuals \cite{birch2024effective}. The conflict initially concentrated in Khartoum state before expanding to western and southern regions, particularly Darfur, where attacks on civilians, healthcare facilities, and essential infrastructure are extensively documented \cite{milton2025counting, dahab2025war}. The impact of conflict on health has been highlighted in recent studies \cite{eljack2023physician, alrawa2023five}.

\begin{figure}[!ht]
\centering
{\includegraphics[width=0.99\linewidth]{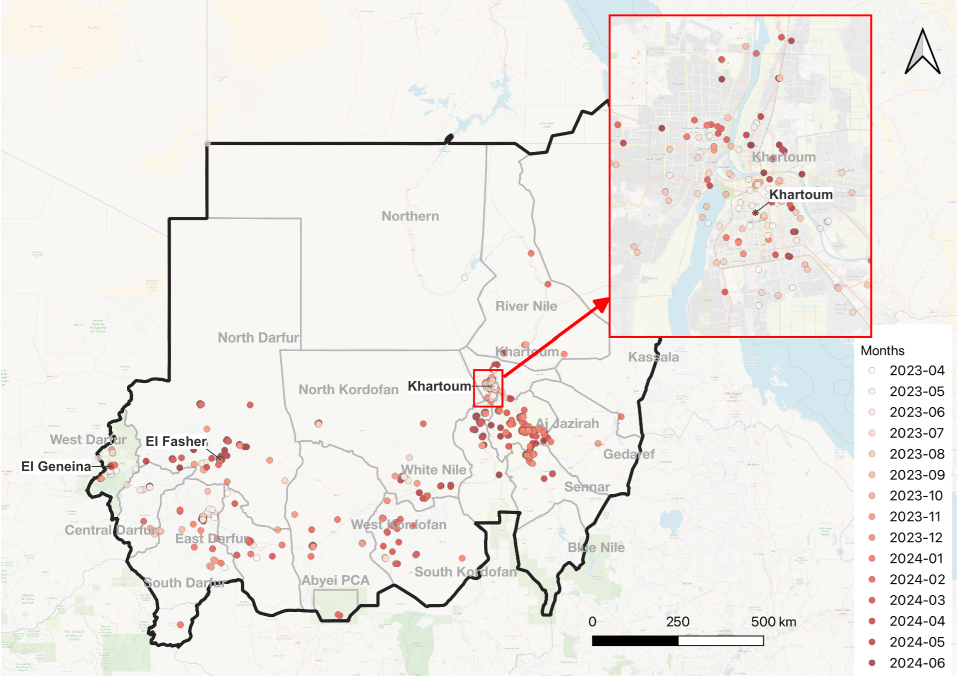}}
\caption{Map of attacks in Sudan, April 2023–June 2024, highlighting the key conflict hotspots.}\label{conflict_intro}
\end{figure}

The Sudan Conflict Observatory (SCO) has reported a massacre of more than a thousand civilians by June 2023 in El-Geneina \cite{rothbart2025sudan}, and reported destruction of healthcare facilities \cite{alma9947703264904105}. Figure \ref{conflict_intro} shows the map of attacks in Sudan. The timeline indicates a rising number of attacks since April 2023, with a significant increase in the regions of Khartoum, El Fasher, and El Geneina.

\subsection{Problem Statement and Objectives}
Monitoring conflict-related fires in active war zones presents several challenges. Ground-based reporting is often delayed, incomplete, or impossible due to insecurity and access constraints \cite{sticher2023toward}. As a result, satellite imagery has become a critical source of independent evidence for tracking attacks, infrastructure destruction, and potential violations of international humanitarian law \cite{hassan2025gender}. Among observable conflict evidence, active fires and burn scars are particularly informative, as they frequently accompany airstrikes, shelling, looting, and the destruction of civilian structures. Recent advances in high-resolution, high-frequency satellite imagery, combined with deep learning techniques, enable the possibility of near–real-time conflict monitoring. Commercial satellite constellations, such as Planet Labs, provide near-daily revisit capabilities at meter-scale resolution, which makes them particularly suitable for detecting small-scale fires in dense environments. Similarly, deep learning methods provide a way to detect anomalous events without relying on ground-truth labels, which are limited in active conflict zones \cite{aung2021using}. However, existing fire-detection and damage-assessment methods face key limitations in conflict settings. Many approaches rely on coarse-resolution sensors, supervised learning pipelines, or computationally intensive models that are not easily scalable for rapid deployment \cite{racek2025unsupervised}. Moreover, models trained on wildfire datasets often perform poorly in urban conflict zones, where fires are short-lived, spatially fragmented \cite{mueller2021monitoring, di2023detection}, and embedded in heterogeneous landscapes \cite{pleniou2025role, franquesa2022assessment}. These constraints hinder timely detection of fire-affected areas, which is critical for humanitarian assessment and operational decision-making.

This study addresses these challenges by leveraging high-frequency, high-resolution PlanetScope imagery within a lightweight, unsupervised deep learning framework. The proposed approach enables scalable, near–real-time monitoring in resource-constrained environments, without relying on ground-truth labels or computationally intensive processing pipelines. The following research questions guide this work:

\begin{itemize}
\item Can conflict-related fire-affected areas be reliably detected in near-real time using high-frequency, high-resolution commercial satellite imagery?
\item How effectively can a lightweight, unsupervised deep learning framework identify fire-affected regions compared to traditional pixel-wise baseline methods?
\item What is the marginal utility of short temporal sequences or additional spectral bands compared to single-snapshot multispectral imagery?
\end{itemize}

The scope of this study is limited to satellite-based observations in Sudan and focuses on the detection of fires and burn scars. Retrospective damage assessment and causal attribution are beyond the scope of this work. The framework is designed for near–real-time monitoring, prioritizing the recall of fire-affected areas while maintaining viable precision. We assume the availability of cloud-free imagery or the application of standard cloud-masking techniques, recognizing that dense cloud cover remains a limitation of optical remote sensing. The specific objectives of this study are to:

\begin{itemize}
\item Adapt a Variational Autoencoder (VAE)-based architecture for 3 m-resolution PlanetScope imagery using four spectral bands.
\item Benchmark the framework against pixel-wise baselines using quantitative performance metrics.
\item Assess the impact of additional temporal and spectral information on detection performance.
\end{itemize}

Collectively, these contributions demonstrate the feasibility of an unsupervised approach for near–real-time detection of conflict-related burn signatures. Furthermore, the study empirically quantifies the marginal benefits of short temporal sequences and additional spectral bands for improving detection performance in conflict settings.

\section{Related Work}
\label{related_work}

\subsection{Deep Learning in Fire Detection}
Convolutional Neural Networks (CNNs) are widely adopted for automating fire detection from satellite imagery by identifying thermal anomalies or spectral signatures of burn scars \cite{toan2019deep, seydi2022fire}. Architectures like U-Net improve pixel-wise segmentation, enabling finer distinctions between active fires and background vegetation \cite{hally2019advances}. More recently, Vision Transformers (ViTs) have demonstrated improved performance in capturing long-range spatial dependencies compared to traditional CNNs \cite{saleh2024forest}. However, these supervised models remain heavily dependent on high-quality, labeled training data, which is scarce in active conflict zones. Moreover, while medium-resolution sensors like Moderate Resolution Imaging Spectroradiometer (MODIS) and Visible Infrared Imaging Radiometer Suite (VIIRS) are effective at thermal sensing, they often struggle to capture small-scale fires in dense or irregular settlements \cite{bromley2010relating, zhao2023tokenized}.

\subsection{Challenges in Conflict Monitoring and Damage Assessment}
Automated satellite-based conflict monitoring has advanced substantially, yet manual interpretation remains prevalent for assessing conflict-related damage \cite{avtar2021remote}. Early studies relied on handcrafted features derived from Very High Resolution (VHR) imagery for post-conflict damage assessment, such as in the Gaza Strip \cite{kahraman2016battle}. Subsequent studies applied supervised and semi-supervised learning approaches to detect damaged infrastructure in Syria using VHR imagery \cite{lee2020assessing, mueller2021monitoring}. While effective, these methods depend on expensive imagery, labeled training data, and retrospective pre- and post-event comparisons, limiting their scalability and timeliness over large geographic extents. Research in Sudan and Darfur has historically utilized MODIS data, which, despite its efficacy in detecting large-scale events, struggles to identify the smaller, dispersed fires typical of irregular warfare \cite{bromley2010relating, cornebise2018witnessing}. Time-series and change detection methods have also been explored in conflict contexts \cite{jenerowicz2010post, braun2018assessment}. 

A recent study on the war in Ukraine shows how multi-sensor satellite data can quantify widespread destruction across urban, agricultural, and industrial landscapes by integrating optical and radar imagery for large-scale conflict-induced change detection \cite{karwowska2026integrating}. A similar work using nighttime thermal observations demonstrates how satellite-derived fire activity can serve as a proxy for frontline dynamics, industrial disruption, and urban combat intensity \cite{dingemanse2026conflict}. Another work on the Gaza Strip demonstrates how high-resolution satellite imagery combined with deep learning can detect explosions and quantify building damage at scale during the 2023–2024 conflict \cite{holail2024time}. The study identifies thousands of missile impact sites and shows a continuous increase in damage over time, with more than half of the built environment affected. Case studies across conflict and disaster settings, including Myanmar \cite{marx2019detecting} and Turkey \cite{kurnaz2020forest}, demonstrate the potential of satellite imagery for fire detection, yet a gap remains in fully unsupervised methods capable of processing high-resolution data streams without labels \cite{xu2022hyperspectral}.

Beyond infrastructure damage, remote sensing has also been used to quantify the environmental and economic consequences of conflict. For instance, a recent study utilizes satellite-derived indicators to estimate war-induced agricultural losses across Ukraine, revealing large-scale and persistent reductions in cultivated land \cite{deininger2025using}. Additionally, multimodal satellite data is used to identify disruptions to ecosystems, flooding events, and the collapse of urban systems during prolonged conflict \cite{xu2024remote}.

These works highlight how remote sensing enables consistent, independent monitoring of conflict impacts in data-scarce environments, particularly where ground access is limited or unsafe. Despite these advances, much of the existing literature focuses on post hoc damage assessment, supervised learning frameworks, or medium-resolution imagery, which limits the detection of transient, small-scale phenomena such as short-lived fires in dense urban environments. Furthermore, while recent conflict-focused studies demonstrate the value of satellite data for documenting the environmental scars of war, fewer approaches address rapid detection using unsupervised methods, as proposed in this study. This gap is particularly evident in rapidly evolving conflicts such as Sudan, where fragmented fire events and limited ground validation require methods that are both scalable and label-independent.

\subsection{Unsupervised Learning and Anomaly Detection}
To mitigate label dependency, recent work has increasingly explored unsupervised anomaly detection, in which models learn the distribution of nominal land cover and identify deviations without requiring explicit annotations \cite{an2015variational}. Variational Autoencoders (VAEs) are frequently employed in this setting due to their ability to learn compact latent representations of complex imagery \cite{kingma2013auto}. While many approaches rely on reconstruction error as an anomaly signal, prior studies have shown that reconstruction-based scores can be unreliable in fully unsupervised settings, particularly for high-capacity models \cite{merrill2020modified}. More recent work has therefore emphasized the use of latent representations for comparing imagery across time, framing anomaly detection as a form of change detection in latent space \cite{bergamasco2022unsupervised}. This approach is particularly relevant in conflict zones, where unprecedented events and limited historical labels make representation-based change detection more robust than reconstruction-based anomaly scoring.

\subsection{Sensor Selection and Data Processing}
Satellite sensor choice determines the trade-off between spatial resolution and temporal frequency. Landsat and Sentinel-2 provide valuable multispectral data for burn scar analysis; however, their revisit intervals of 5–10 days often miss short-lived fire events \cite{schroeder2016active}. Synthetic Aperture Radar (SAR) can penetrate cloud cover to assess post-fire land changes, though it lacks the spectral signatures provided by optical sensors \cite{narvaez2023burnt}. PlanetScope imagery offers a balance, providing 3 m resolution and near-daily revisit cadences. However, leveraging this data for conflict monitoring requires robust preprocessing and computationally efficient models capable of handling large daily data volumes, factors that have limited its adoption in fully unsupervised, near–real-time fire detection workflows \cite{boroujeni2024comprehensive, de2021active, bazi2021vision}.

\section{Study Area and Data}
\label{sec2}

\subsection{Study Area}
This study focuses on five conflict-related fire incidents in the El Fasher and Khartoum regions of Sudan. In Khartoum State, an airstrike reportedly carried out by the Sudanese Armed Forces (SAF) on May 25, 2024, targeted the Gandahar Market in Omdurman. The Conflict Observatory documented damage to at least 31 structures resulting from this incident \cite{etefa2026fall}. PlanetScope imagery acquired on the day of the strike shows a visible smoke plume over the market area, consistent with active burning.

In North Darfur, political and security tensions escalated in late March 2024, with both the SAF and the Rapid Support Forces (RSF) preparing for confrontation in El Fasher by early May. According to the Conflict Observatory, at least 7,800 structures in eastern El Fasher were affected by fighting beginning on May 9, 2024, with NASA FIRMS data indicating widespread fire activity and extensive burning by May 11 \cite{alma9947677646204105}. Earlier, in April 2024, RSF-led forces attacked approximately 17 villages surrounding El Fasher. These attacks caused significant civilian casualties and widespread destruction of residential structures \cite{sabahelzain2025fasher}. PlanetScope imagery captured active fires in several of these locations on April 13, 2024, including the villages of Jaranga, Muqrin, and Sarafaya \cite{alma9947677655704105}. 

\subsection{Imagery details}
Planet operates Earth-imaging constellations: PlanetScope (PS), RapidEye (RE), and SkySat (SS). Imagery formats serve various use cases, including deep learning, disaster response, precision agriculture, and temporal image analytics. Planet offers three product lines for PlanetScope imagery: Basic Scene, Ortho Scene, and Ortho Tile. The Basic Scene product is a scaled Top of Atmosphere Radiance (at sensor) and sensor-corrected product, suitable for advanced image processing and geometric correction. Ortho Tiles are multiple orthorectified scenes in a single strip, divided according to a defined grid. Ortho Scenes represent single-frame image captures with post-processing, removing terrain distortions and suitable for cartographic purposes. Ortho Scenes are delivered as visual (RGB) and analytic products \cite{marta2018planet}. In this work, we utilized radiometrically calibrated 4-band and 8-band ortho-analytic imagery obtained from the PSB.SD instrument. The instrument is built with the ``PSBlue’’ telescope with a 47-megapixel sensor. It captures 8 bands including red, green, blue, near infrared, red edge, green I, coastal blue, and yellow channels. The butcher-block filter of the instrument consists of four individual pass-band filters, which separate light into the blue, green, red, and NIR channels. The pass-band filters allow interoperability with 6 bands of Sentinel-2. It produces scene products which are approximately $32.5 \times 19.6$ $km^2$ \cite{planetdocs}.

\subsection{Incident data}
The SCO methodology for collecting incident data uses a multi-phase approach combining secondary research and original data collection. Researchers first examine reports from organizations including Human Rights Watch, the World Health Organization, and the United Nations. Additionally, Armed Conflict Location \& Event Data (ACLED) catalog is utilized as another complementary source to consolidate the incidents dataset \cite{carboni2024collecting}. These sources document civilian targeting and infrastructure destruction, which the SCO then aims to enhance with further analysis or more current accounts from the impacted areas in Sudan. Social media provides direct witness reports and real-time updates for incident verification. By combining these varied streams, the SCO researchers confirm the validity of specific incidents. Furthermore, Sudan Human Rights Hub (SHRH) conducted interviews of individuals on the ground affected by the attacks. Testimony was given only after the research participants received full information about the purpose, scope and use of their information, followed by their consent. Detailed methodology and technical specifications of the incident collection methodology are reported in \cite{masri2025automating}.

SCO reports cover a wide range of events, including infrastructure damage, civil rights violations, and gender-based violence. For this study, we focused on incidents attributed to fire, as these produce visual signatures that can be captured and validated in satellite imagery. To ensure suitable data for evaluation, incidents were filtered to include only those with cloud-free, accurately co-registered PlanetScope imagery acquired within 24 hours of the reported event (Table \ref{imagery_details}). Leveraging the near-daily revisit capability of PlanetScope, three consecutive pre-incident images (Pre-1 to Pre-3) were selected at approximately 1-day intervals when available, enabling assessment of short-term temporal variability. For the single pre-incident image, the closest cloud-free, radiometrically consistent acquisition (Pre-3) was used. This choice minimizes unrelated land surface variability, consistent with remote sensing change detection practices favoring shorter temporal intervals to reduce confounding effects such as seasonal variation and illumination differences \cite{coppin1996digital, verbesselt2010detecting}.

Temporal evaluation was conducted by measuring the delay $\Delta t = t_{\text{post}} - t_{\text{incident}}$ between the reported incident time and the first available cloud-free post-incident acquisition, which serves as the earliest detection opportunity under satellite revisit constraints. A detection is considered timely if fire-affected areas are identified in this first cloud-free post-incident image (typically within 24–30 hours). Missed detections are defined as cases where no anomaly is observed in the first available post-incident acquisition, and the corresponding delay is treated as right-censored at the time of the subsequent observation.

\begin{table}[h!]
\tiny
\caption{Details of satellite imagery used for the study area}
\label{imagery_details}
\centering
\begin{tabular}{ccccccc}
\toprule
\textbf{Place} & \textbf{Area} & \textbf{Incident} & \textbf{Pre-1} & \textbf{Pre-2} & \textbf{Pre-3} & \textbf{Post} \\
 & \textbf{(\(km^2\))} & \textbf{date} & \textbf{date} & \textbf{date} & \textbf{date} & \textbf{date} \\
\midrule
Gandahar Market & 4.37 & May 25, 2024 & May 11, 2024 & May 13, 2024 & May 14, 2024 & May 25, 2024 \\
El Fasher & 5.41 & May 09--11, 2024 & Apr 29, 2024 & May 01, 2024 & May 02, 2024 & May 11, 2024 \\
Muqrin & 4.52 & Apr 13, 2024 & Feb 27, 2024 & Feb 28, 2024 & Mar 01, 2024 & Apr 13, 2024 \\
Jaranga & 2.31 & Apr 13, 2024 & Feb 27, 2024 & Feb 28, 2024 & Mar 01, 2024 & Apr 13, 2024 \\
Sarafaya & 2.16 & Apr 13, 2024 & Feb 27, 2024 & Feb 28, 2024 & Mar 01, 2024 & Apr 13, 2024 \\
\bottomrule        
\end{tabular} 
\end{table}

\begin{figure}[!ht]
\centering
{\includegraphics[width=0.99\linewidth]{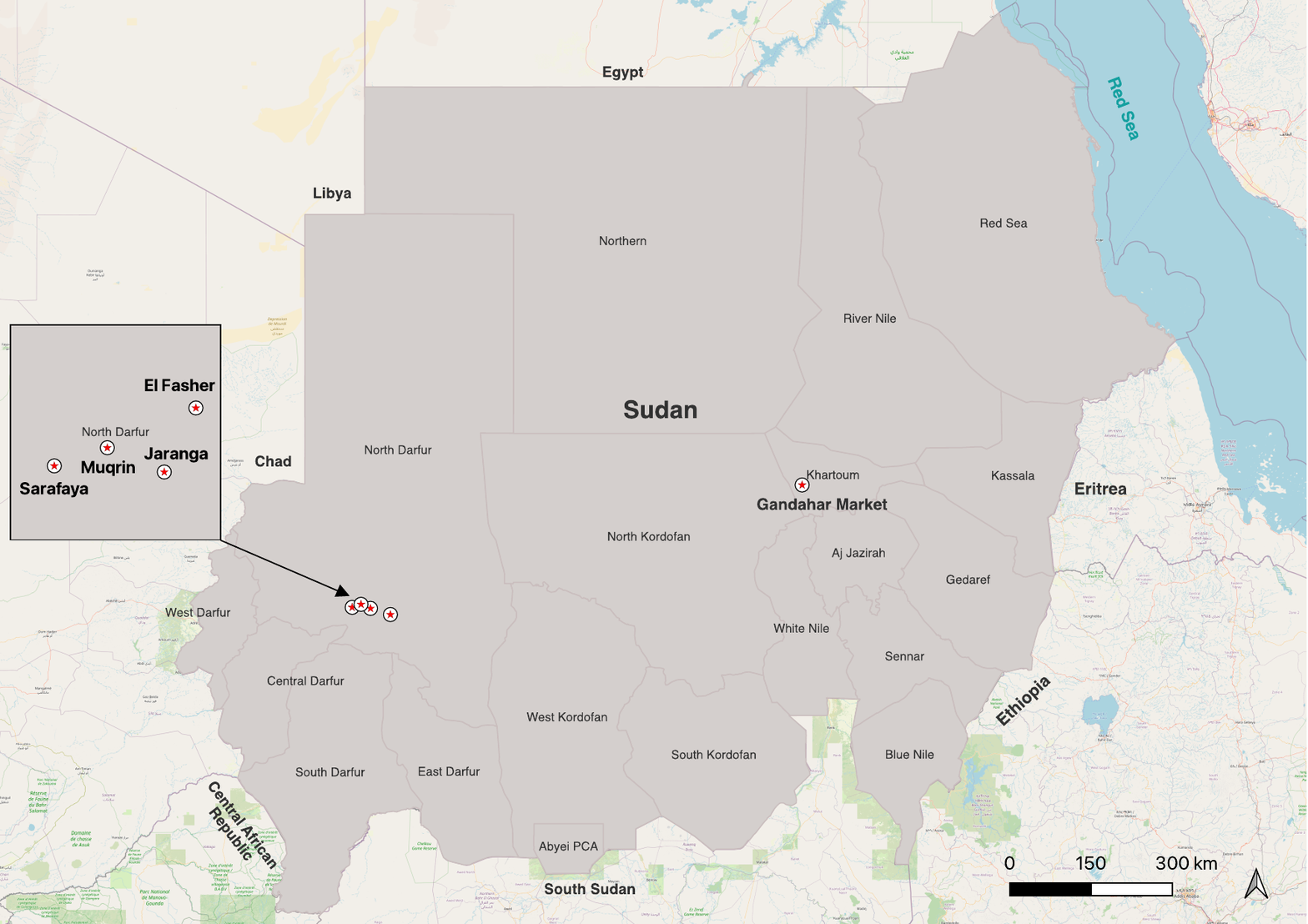}}
\caption{The depiction of study areas used for this research.}\label{study_area}
\end{figure}

Table \ref{imagery_details} summarizes the imagery and incident details for all case studies considered in this work. Additionally, Figure \ref{study_area} illustrates the geographic distribution of the study areas across Sudan. These incidents were selected based on independently reported fire-related damage documented in \cite{yhrlarson, sabahelzain2025fasher, yhrlelfasher}, and manual evaluation and verification conducted by analysts associated with the Sudan Conflict Observatory project.

\section{Methodology}
\label{sec3}

\subsection{Model architecture}

In this study, we adapted and retrained RaVAEn \cite{ruuvzivcka2022ravaen}, a lightweight convolutional Variational Autoencoder (VAE) framework originally developed for unsupervised detection of disaster-related surface anomalies across multiple hazard types, including floods, fire burn scars, landslides, and hurricanes. Rather than being tailored to a specific phenomenon, RaVAEn learns a compact latent representation of nominal surface conditions and identifies anomalous patterns as deviations from this learned distribution, making it suitable for detecting previously unseen or irregular events such as conflict-related fires. The original model was designed for 10-band satellite imagery at 10 m spatial resolution. We retrained it for 3 m resolution imagery with four spectral channels (RGB and near-infrared) to better capture fine-scale spatial heterogeneity associated with small, fragmented fire events in urban conflict settings.

The revised model employs a resolution-adaptive Variational Autoencoder (VAE) designed to extract scale-robust surface characteristics from heterogeneous satellite imagery while reducing domain discrepancies across varying ground sampling distances. The model follows an unsupervised encoder–decoder framework composed of four hierarchical stages, with adaptations concentrated in the early layers to address resolution-dependent spatial frequencies present in 10 m and 3 m data. At the input, the encoder applies a frequency decomposition layer, where a fixed Gaussian low-pass filter extracts coarse structural priors and a high-pass residual captures fine-grained texture. These components are concatenated to provide the encoder with explicit access to both coarse structure and fine detail, enabling consistent feature learning across resolutions.

\begin{figure}[t]
\centering
\resizebox{\textwidth}{!}{%
\begin{tikzpicture}[
    font=\small,
    node distance=0.8cm and 1cm,
    every node/.style={align=center},
    block/.style={draw, trapezium, trapezium left angle=70, trapezium right angle=110, minimum height=1.5cm, minimum width=2.2cm},
    rect/.style={draw, rectangle, minimum height=1.2cm, minimum width=2cm, rounded corners=1pt},
    smallrect/.style={draw, rectangle, minimum height=0.8cm, minimum width=0.8cm, rounded corners=1pt},
    arrow/.style={->, thick, rounded corners=3pt}
]

\node (input) {Input Tile\\$32\times32\times4$};

\node[rect, right=of input] (freq) {Frequency\\Decomposition\\\footnotesize (Low + High)};
\node[rect, right=of freq] (concat) {Concatenation};
\node[rect, right=of concat] (msconv) {Multi-scale\\Conv Block\\\footnotesize $d=1,2,4$};
\node[rect, right=of msconv] (blur) {BlurPool};

\node[rect, below=1.8cm of input] (res) {Residual\\Blocks\\\footnotesize (LeakyReLU + Norm)};
\node[rect, right=of res] (pool) {Global\\Pooling};

\node[smallrect, right=1.2cm of pool, yshift=0.5cm] (sigma) {$\sigma$};
\node[smallrect, right=1.2cm of pool, yshift=-0.5cm] (mu) {$\mu$};

\path (sigma) -- (mu) coordinate[midway] (musigmid);
\node[smallrect, right=1cm of musigmid] (z) {$z \in \mathbb{R}^{128}$};

\node[block, right=of z] (decoder) {Decoder\\$p_\theta(x|z)$};
\node[right=of decoder] (out) {Reconstruction\\$32\times32\times4$};

\draw[arrow] (input) -- (freq);
\draw[arrow] (freq) -- (concat);
\draw[arrow] (concat) -- (msconv);
\draw[arrow] (msconv) -- (blur);

\draw[arrow] (blur.south) -- ++(0,-0.6cm) -| (res.north);

\draw[arrow] (res) -- (pool);

\draw[arrow] (pool.east) -- ++(0.4,0) |- (sigma.west);
\draw[arrow] (pool.east) -- ++(0.4,0) |- (mu.west);

\draw[arrow] (sigma.east) -- ++(0.4,0) |- (z.west);
\draw[arrow] (mu.east) -- ++(0.4,0) |- (z.west);

\draw[arrow] (z) -- (decoder);
\draw[arrow] (decoder) -- (out);

\node[above=0.2cm of z] (zlabel) {\footnotesize $z \sim \mathcal{N}(\mu, \sigma^2)$};

\node[draw, dotted, inner sep=0.4cm, fit=(input)(freq)(concat)(msconv)(blur)(res)(pool)(sigma)(mu)(z)(zlabel)(decoder)(out)] {};

\end{tikzpicture}
}
\caption{A schematic diagram of the VAE architecture used in this study.}
\label{model_arch}
\end{figure}

To accommodate scale-variant spatial patterns, standard strided convolutions in the initial stages are replaced with multi-scale convolutional blocks analogous to Atrous Spatial Pyramid structures \cite{maggiori2016convolutional}. These blocks use parallel $3\times3$ convolutions with increasing dilation rates, fused via a $1\times1$ convolution. This design enables the encoder to capture multi-scale contextual dependencies and expand its effective receptive field while preserving spatial detail \cite{chen2017rethinking, yuan2018multiscale}. Anti-aliased downsampling, or BlurPool \cite{zhang2019making}, implemented as a stride-1 convolution followed by low-pass filtering and pooling, mitigates the loss of high-frequency information and improves shift equivariance when processing higher-resolution imagery. Each downsampling operation is followed by residual convolutional blocks with leaky ReLU activations and normalization layers, while deeper stages retain a conventional residual structure as higher-level semantic features are less sensitive to pixel-level resolution.

The final encoder feature map is aggregated using global pooling and projected onto a 128-dimensional latent space parameterized by a diagonal Gaussian distribution, regularized with a $\beta$-VAE objective ($\beta$=1). The decoder mirrors the encoder hierarchy using nearest-neighbor upsampling and convolutional layers, with a linear activation in the output layer. The model is trained end-to-end using the Adam optimizer with a learning rate of 0.001 and a batch size of 128, minimizing a combination of mean squared reconstruction loss and Kullback–Leibler divergence. To improve generalization across spatial resolutions, a scale-augmented training strategy is applied in which imagery is stochastically resampled to simulate continuous transitions in ground sampling distance \cite{koziarski2017image}. This resampling is applied only during training; inference operates directly on native-resolution imagery, producing resolution-invariant feature embeddings. Figure \ref{model_arch} illustrates the model architecture.

While VAEs are commonly used for unsupervised anomaly detection by thresholding reconstruction error for individual inputs \cite{angerhausen2022unsupervised}, prior work has shown that reconstruction error can be an unreliable indicator of anomalous behavior in fully unsupervised settings \cite{merrill2020modified}. Motivated by these findings, we do not use reconstruction error as the anomaly signal. Instead, the VAE is used as a representation learning framework to embed surface conditions into a compact latent space, in which changes between temporally paired observations can be quantified for change detection.

\subsection{Preprocessing and Spectral Alignment}

To mitigate sensor-specific radiometric bias and atmospheric variability, we implement a dual-stage normalization pipeline. Initially, to align the 3-meter imagery with the 10-meter reference, we perform Spectral Normalization via Major Axis (MA) Linear Regression \cite{chastain2019empirical}. For each overlapping spectral band $b$, the calibrated 3-meter reflectance $x_{\text{cal}}$ is derived as:

\begin{equation}
x_{\text{cal}, b} = \alpha_b \cdot x_{\text{3m}, b} + \beta_b
\end{equation}

where the gain ($\alpha$) and offset ($\beta$) coefficients are empirically derived from near-coincident acquisitions. Following spectral alignment, we address the high dynamic range and non-linear distribution of the near-infrared (NIR) band relative to visible spectra. To ensure balanced feature importance across the multispectral cube, we apply a logarithmic compression followed by robust min-max feature scaling. 

Consistent with the original RaVAEn configuration, imagery is partitioned into tiles of $32 \times 32$ pixels. Each tile is normalized to the $[-1, +1]$ interval according to:

\begin{equation}
x' = \log(x_{\text{cal}} + \epsilon), \quad x'' = 2 \cdot \frac{x' - P_{1}(x')}{P_{99}(x') - P_{1}(x')} - 1
\end{equation}

where $\epsilon$ is a small constant to ensure numerical stability and $P_{1}, P_{99}$ represent the $1^{\text{st}}$ and $99^{\text{th}}$ percentiles computed per band across the training distribution. This robust scaling minimizes sensitivity to sensor artifacts and extreme outliers while preserving the variance of surface reflectance characteristics. The resulting transformation parameters are archived and consistently applied during inference to maintain latent space stability.

\subsection{Training and Inference}
\textbf{Training:} The model was trained using the WorldFloods dataset \cite{mateo2021towards} derived from Sentinel-2 multi-spectral imagery \cite{fletcher2012sentinel}. The dataset includes four classes of disaster-related surface changes, including fire burn scars. The dataset was selected for its high variance in spectral and textural patterns. Training on diverse disaster-related surface changes encourages the VAE to learn generalized spatial and structural features rather than overfitting to specific biome characteristics. This capability enables the model to effectively reconstruct nominal arid backgrounds in Sudan and highlight fire-related deviations as anomalies, despite the ecological domain shift. Training was performed on 233 scenes, with a held-out validation set of 19 scenes used to assess convergence and prevent overfitting. The anomalous events in the validation set were identified using the Copernicus EMS system. The validation images with greater than 20\% cloud cover were discarded to mitigate the effects of cloud cover. Training and validation were conducted on a machine equipped with one NVIDIA H100 GPU, 112 Intel CPU cores, and 2 TB of system memory.

We trained the model for 200 epochs based on the convergence analysis of the VAE objective on a held-out validation set consisting of 19 scenes. The reconstruction loss decreased rapidly during early training and exhibited diminishing improvements after approximately 80 epochs, while the Kullback–Leibler divergence term stabilized after around 120 epochs, indicating convergence of the latent distribution. Beyond 200 epochs, the total validation loss plateaued and no further gains were observed in downstream change detection performance, suggesting that additional training yielded negligible benefit. Therefore, 200 epochs provide sufficient optimization for stable convergence without incurring unnecessary computational cost.

\textbf{Inference:} For conflict monitoring in Sudan, radiometrically calibrated 4-band ortho-analytic imagery with red, green, blue, and near-infrared bands was partitioned into $32 \times 32$ pixel tiles. Corresponding pre- and post-incident tiles from the same spatial locations were passed through the encoder to obtain latent embeddings. Anomaly detection is formulated as change detection in latent space rather than reconstruction-based scoring. Anomaly scores were computed using the cosine distance between the latent representations of each tile pair, capturing deviations from nominal surface conditions learned during training. This representation-based comparison reduces sensitivity to absolute radiometric differences and sensor noise, facilitating the sensitive detection of fire-affected areas without requiring labeled training data. Inference was performed on 5 scenes, one for each of the case studies.

\subsection{Evaluation and Metrics}
\label{methodology_eval}
\textbf{Comparison with conventional methods:} 
We conducted a comparative evaluation using standard pixel-level change detection approaches, including cosine distance, Canonical Variates Analysis (CVA) \cite{malila1980change}, and Iteratively Reweighted Multivariate Alteration Detection (IR-MAD) \cite{nielsen2007regularized}. Cosine distance was computed between corresponding pre- and post-incident tiles to produce continuous anomaly scores based on spectral dissimilarity. CVA captures change in the original spectral feature space, while IR-MAD provides statistically derived change significance measures that emphasize invariant background structure.

To further assess task-specific performance for fire detection, we incorporated two directional spectral baselines: the Differenced Normalized Difference Vegetation Index (dNDVI) and the Differenced Burn Area Index (dBAI). While the VAE models general spectral dissimilarity, the dNDVI baseline was formulated as a directional measure of vegetation loss, computed as $\mathrm{NDVI}_{\text{pre}}-\mathrm{NDVI}_{\text{post}}$, such that higher values indicate greater reductions in green biomass. In contrast, dBAI was employed as a measure of charcoal accumulation between pre- and post-incident tiles, with higher values indicating a spectral shift toward the charcoal reference endmember. These transformations align all methods with a consistent ``higher score indicates greater anomaly'' convention while preserving their physical interpretability as task-aligned benchmarks.

By holding preprocessing, tile geometry, and temporal pairing constant across all methods, differences in performance isolate the contribution of the VAE-based latent representations, independent of heuristic thresholds. Scores from comparison methods were thresholded at the 95th percentile of their respective score distributions, enabling direct comparison with VAE-based anomaly scores derived from latent embeddings.

\textbf{Fire incident labeling:}
Ground-truth labels were curated as part of the SCO initiative, which analyzes information collected from human rights organizations, ACLED, and field interviews \cite{etefa2026fall, alma9947677646204105}. Tiles were labeled fire-affected if post-incident imagery revealed burn scars or active fire. Each scene was reviewed by at least three analysts, with discrepancies resolved by consensus, yielding inter-annotator agreement above 90\%. To mitigate circularity and spatial uncertainly, fire incidents were further validated using FIRMS and VIIRS datasets and cross-referenced with independent reports \cite{yhrlarson, yhrlelfasher}. Labels were used solely for retrospective evaluation and not for training, threshold optimization, or model selection. Visual interpretation of PlanetScope imagery was limited to the spatial delineation of burn scars for independently reported and verified incidents. This approach ensures that the model is evaluated on its ability to segment confirmed damage, rather than on incidents identified directly from satellite imagery.

\textbf{Metrics:} 
Model performance was quantified using the area under the precision–recall curve (AUPRC), which provides a threshold-independent measure of how effectively anomaly scores rank fire-affected tiles above non-fire tiles. AUPRC is preferred over the ROC curve in this setting because fire-affected tiles constitute a small fraction of each scene, making precision–recall curves more informative under severe class imbalance. Continuous anomaly scores were derived from cosine distances in the VAE latent space and from pixel-space distance or transformation-based measures for the comparison methods, and were evaluated against curated fire labels. Precision and recall were computed by sweeping thresholds across all observed scores to generate binary predictions, producing a precision–recall curve summarizing the trade-off between true and false positives. Representative thresholds were selected to compute precision, recall, and F1-score, reported as median point estimates across folds, including commission and omission error rates. Beyond quantitative metrics, model outputs were assessed qualitatively by visualizing the spatial distribution of anomalous tiles, allowing inspection of spatial coherence and alignment with observed fire patterns.

\textbf{Near–real-time detection:} 
We define ``near–real-time'' detection as the potential to produce fire anomaly outputs within approximately 24–30 hours of image acquisition under favorable imaging conditions. This definition leverages the near-daily revisit cadence of the PlanetScope constellation, which consists of approximately 130 CubeSat satellites capable of imaging the Earth’s land surface on a daily basis \cite{marta2018planet}, providing high temporal coverage suitable for rapid conflict monitoring. Under nominal conditions, this latency includes image acquisition and delivery (approximately 18–20 hours for PlanetScope imagery), preprocessing and tiling (spectral standardization, logarithmic scaling, and $32 \times 32$ pixel tile generation in 1–2 hours), and model inference plus postprocessing (latent-space scoring, anomaly mapping, and visualization in approximately 1–2 hours per scene on GPU-enabled hardware). While these stages suggest that rapid situational awareness may be feasible under favorable conditions, actual operational latency may be longer in conflict zones due to cloud cover, image misregistration, platform-specific delivery policies, or tasking constraints, which were not explicitly stress-tested in this study. Despite the near-daily imaging capability, not all acquired imagery can be operationally utilized. A small proportion of scenes cannot be fully processed due to image acquisition anomalies, atmospheric contamination, or other quality-related issues. In addition, accurate georeferencing depends on successful ground lock, and approximately 95\% of ground-lock failures are associated with cloud cover \cite{planetdocs}. Consequently, image availability in conflict zones may be intermittent despite frequent satellite overpasses. Therefore, this definition should be interpreted as an ideal benchmark under cloud-free and well-registered imaging conditions, while real-world deployment may require additional buffering to account for image availability and quality limitations.

\section{Results}
\label{sec4}

We evaluated the effectiveness of the proposed deep learning–based fire detection approach by comparing model outputs against both general-purpose and task-specific change detection baselines, including cosine distance, CVA, IR-MAD, and two spectral indices: dNDVI and dBAI. Performance was assessed across five case studies in Sudan using qualitative visual inspection of prediction maps and quantitative metrics. To quantify uncertainty and robustness, we estimated 95\% confidence intervals (CIs) for all performance metrics using 1,000 bootstrap resamples of tiles within each scene. Additionally, paired Wilcoxon signed-rank tests were applied to AUPRC values obtained from identical bootstrap resamples of tiles across methods, ensuring matched, nonparametric comparison under identical sampling conditions.

Across the five case studies, the VAE-based approach exhibits varying levels of sensitivity and robustness depending on scene characteristics. In scenarios with strong burn signals, the model achieves high recall and clearly delineates fire-affected areas, while in high-noise or degraded conditions, spectral and structural ambiguity can reduce accuracy. Spectral index baselines exhibit performance that is highly dependent on scene composition: dNDVI is generally less effective in sparsely vegetated or urban environments, while dBAI provides more consistent detection of burn signatures but remains sensitive to confounding materials such as dark rooftops, soil, and shadows.

\begin{figure}[h!]
    \centering
    \begin{subfigure}{0.32\textwidth}
        \centering
        \includegraphics[width=\linewidth]{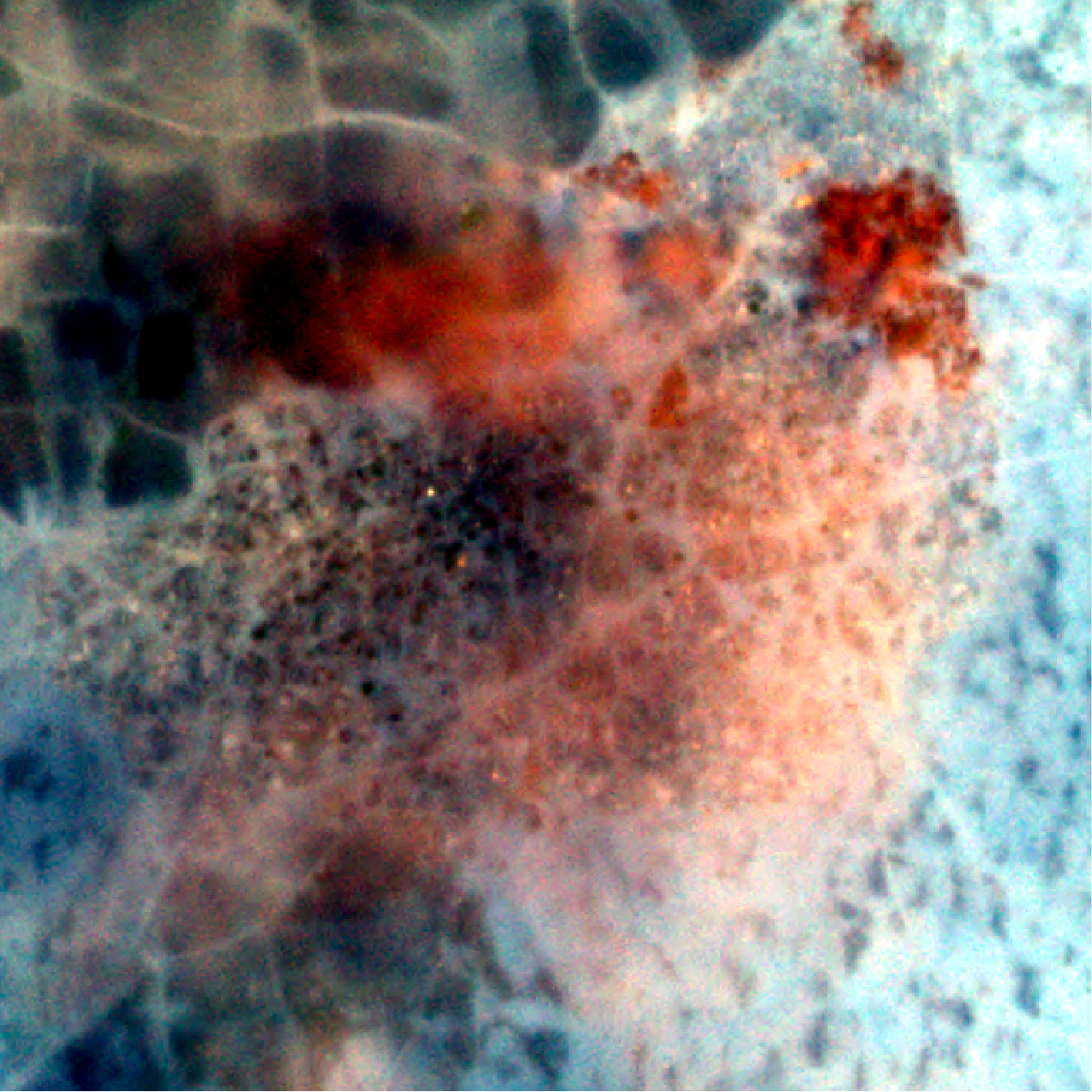}
        \caption{Input}
    \end{subfigure}
    \hfill
    \begin{subfigure}{0.32\textwidth}
        \centering
        \includegraphics[width=\linewidth]{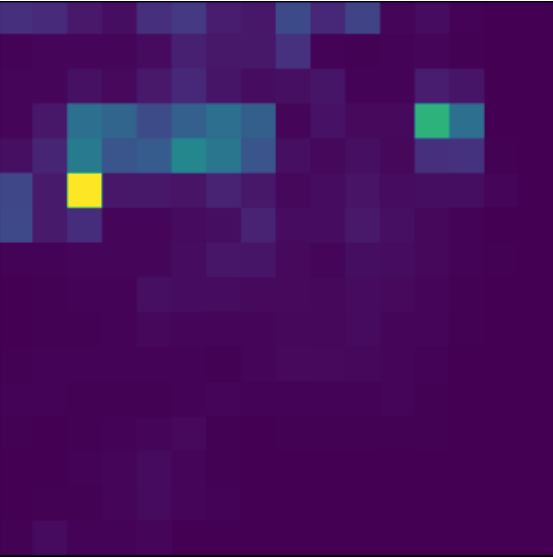}
        \caption{Cosine distance}
    \end{subfigure}
    \hfill
    \begin{subfigure}{0.32\textwidth}
        \centering
        \includegraphics[width=\linewidth]{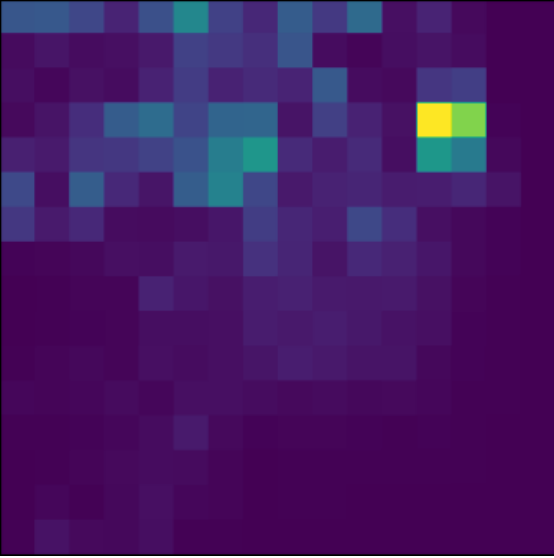}
        \caption{Prediction}
    \end{subfigure}
    \caption{Comparison of cosine distance with prediction in Jaranga.}\label{jaranga_base}
\end{figure}

\begin{figure}[h!]
    \centering
    \begin{subfigure}{0.32\textwidth}
        \centering
        \includegraphics[width=\linewidth]{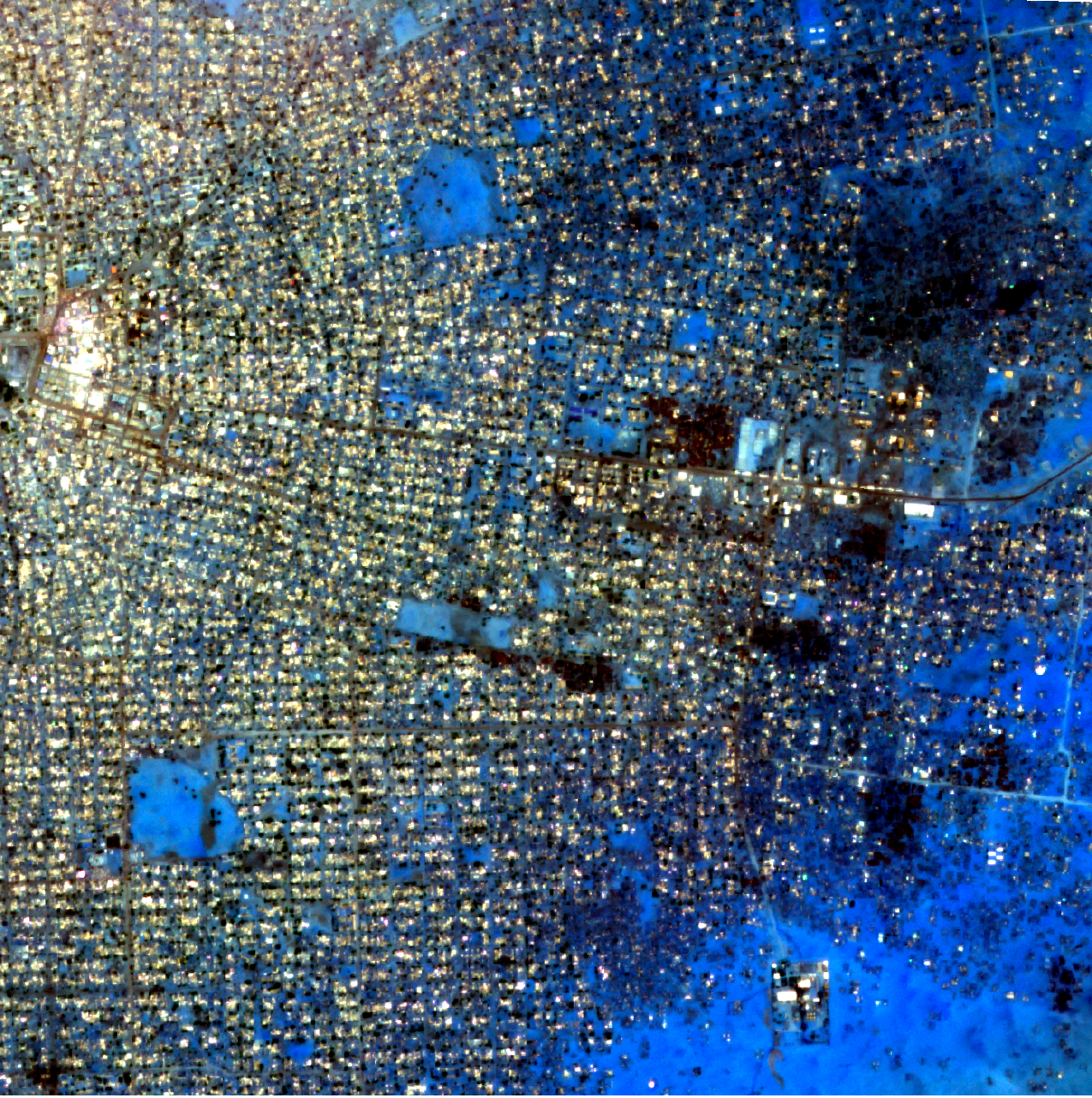}
        \caption{Input}
    \end{subfigure}
    \hfill
    \begin{subfigure}{0.32\textwidth}
        \centering
        \includegraphics[width=\linewidth]{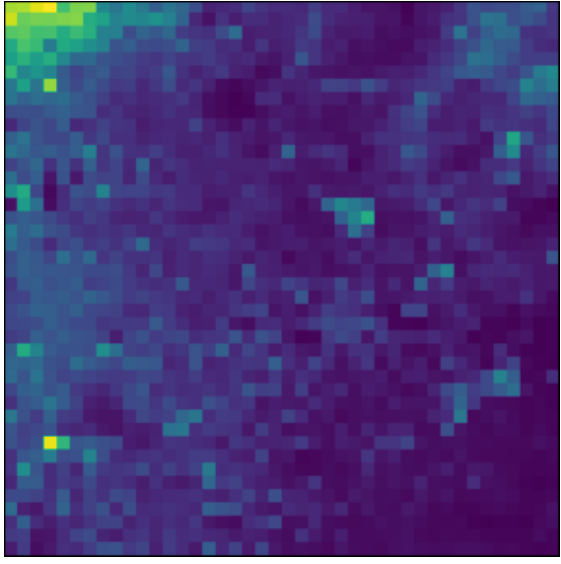}
        \caption{Cosine distance}
    \end{subfigure}
    \hfill
    \begin{subfigure}{0.32\textwidth}
        \centering
        \includegraphics[width=\linewidth]{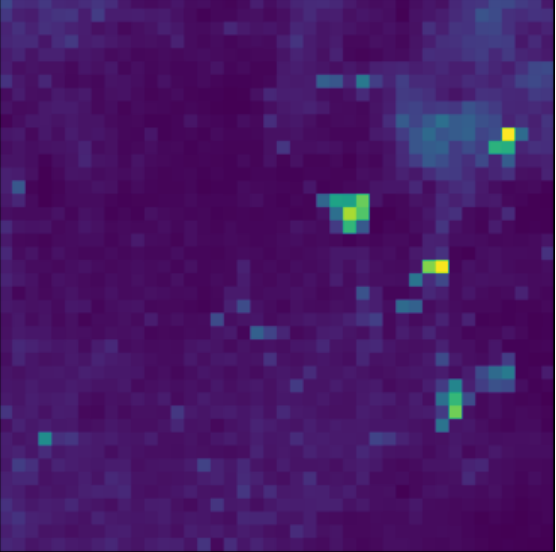}
        \caption{Prediction}
    \end{subfigure}
    \caption{Comparison of cosine distance with prediction in El Fasher.}\label{el_fasher_base}
\end{figure}

\begin{figure}[h!]
    \centering
    \begin{subfigure}{0.32\textwidth}
        \centering
        \includegraphics[width=\linewidth]{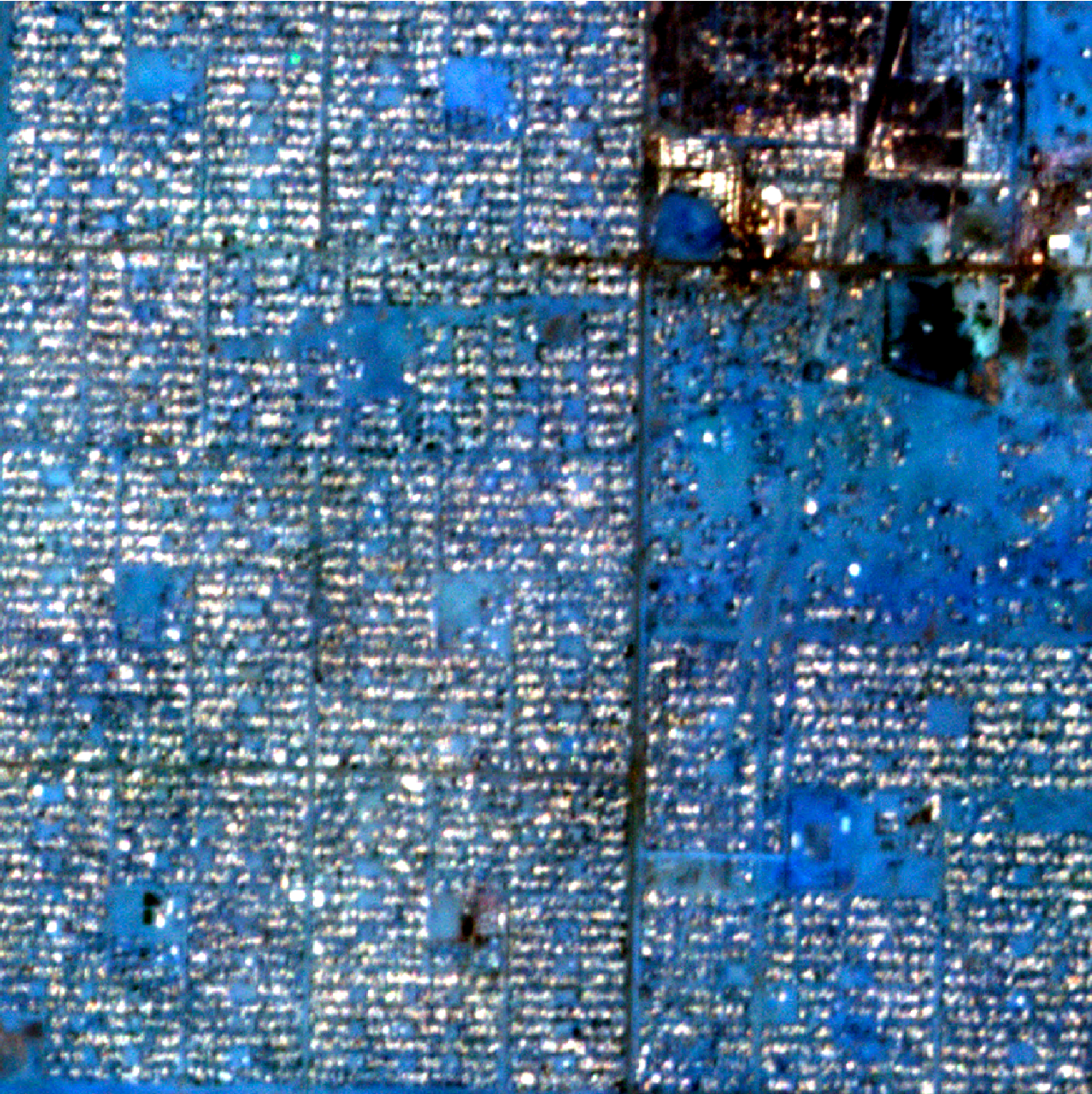}
        \caption{Before}
    \end{subfigure}
    \hfill
    \begin{subfigure}{0.32\textwidth}
        \centering
        \includegraphics[width=\linewidth]{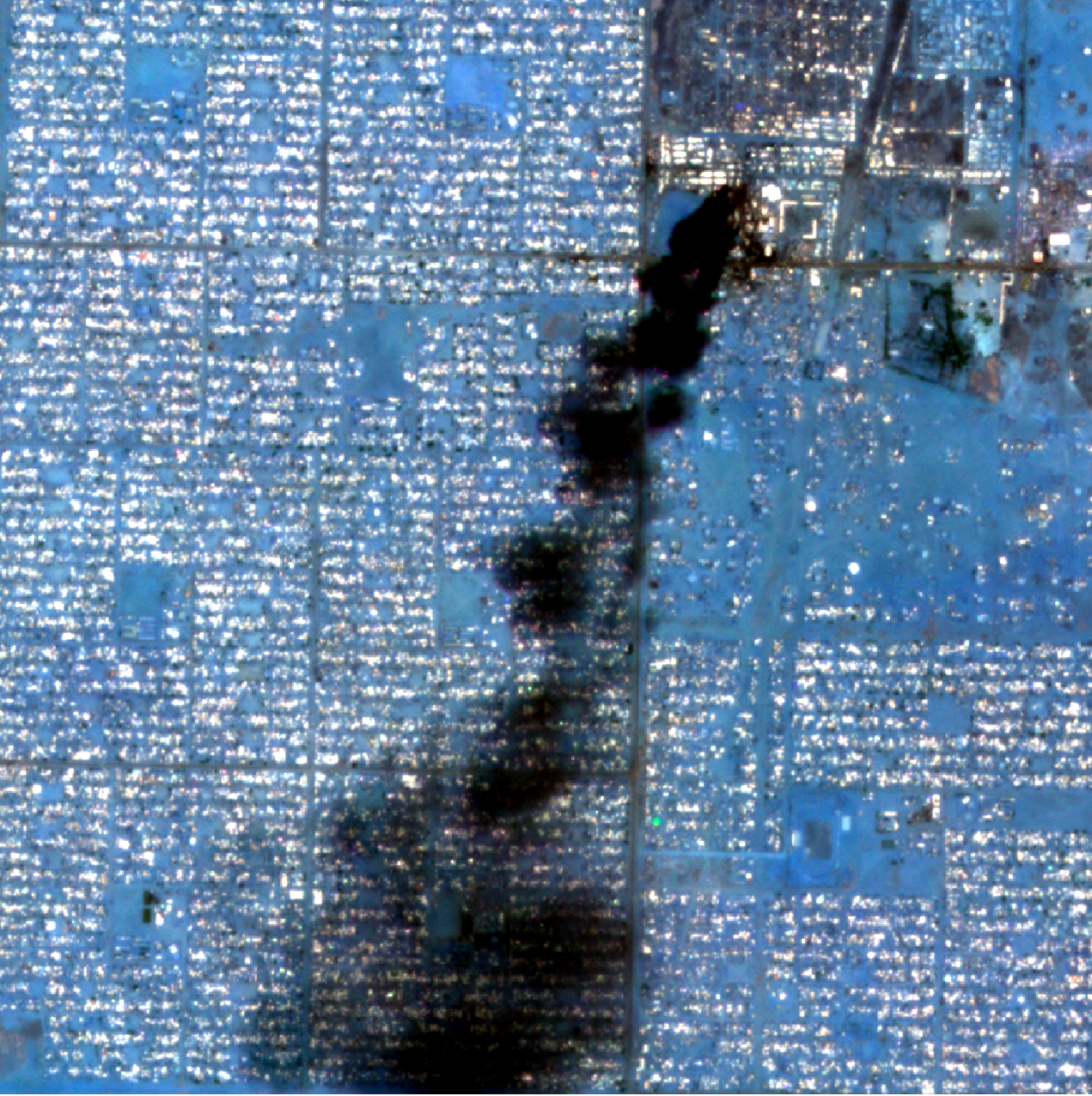}
        \caption{After}
    \end{subfigure}
    \hfill
    \begin{subfigure}{0.32\textwidth}
        \centering
        \includegraphics[width=\linewidth]{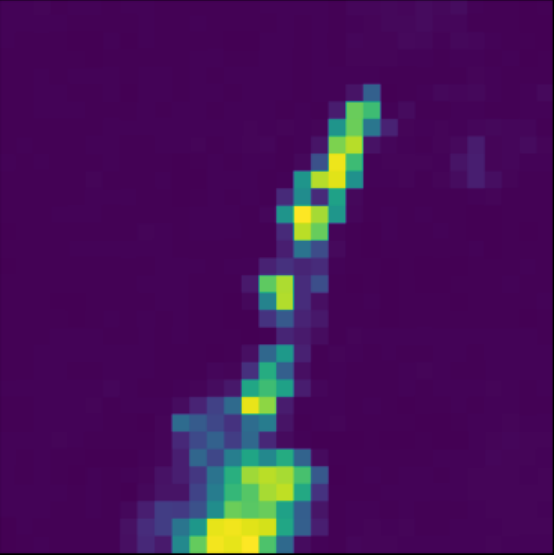}
        \caption{Prediction}
    \end{subfigure}
    \caption{Comparison of before and after input imagery with an active fire prediction in Gandahar Market.}
    \label{gandahar_mkt}
\end{figure}

Figures \ref{jaranga_base} and \ref{el_fasher_base} present visual comparisons with cosine distance for Jaranga and El Fasher, respectively. Results indicate that cosine distance preserves fine-grained details from the input imagery but is sensitive to noise, misregistration, and minor radiometric or geometric variations. In contrast, the proposed VAE-based approach captures higher-level latent features, facilitating the delineation of fire-affected areas relative to background changes.

\begin{figure}[h!]
    \centering
    \begin{subfigure}{0.32\textwidth}
        \centering
        \includegraphics[width=\linewidth]{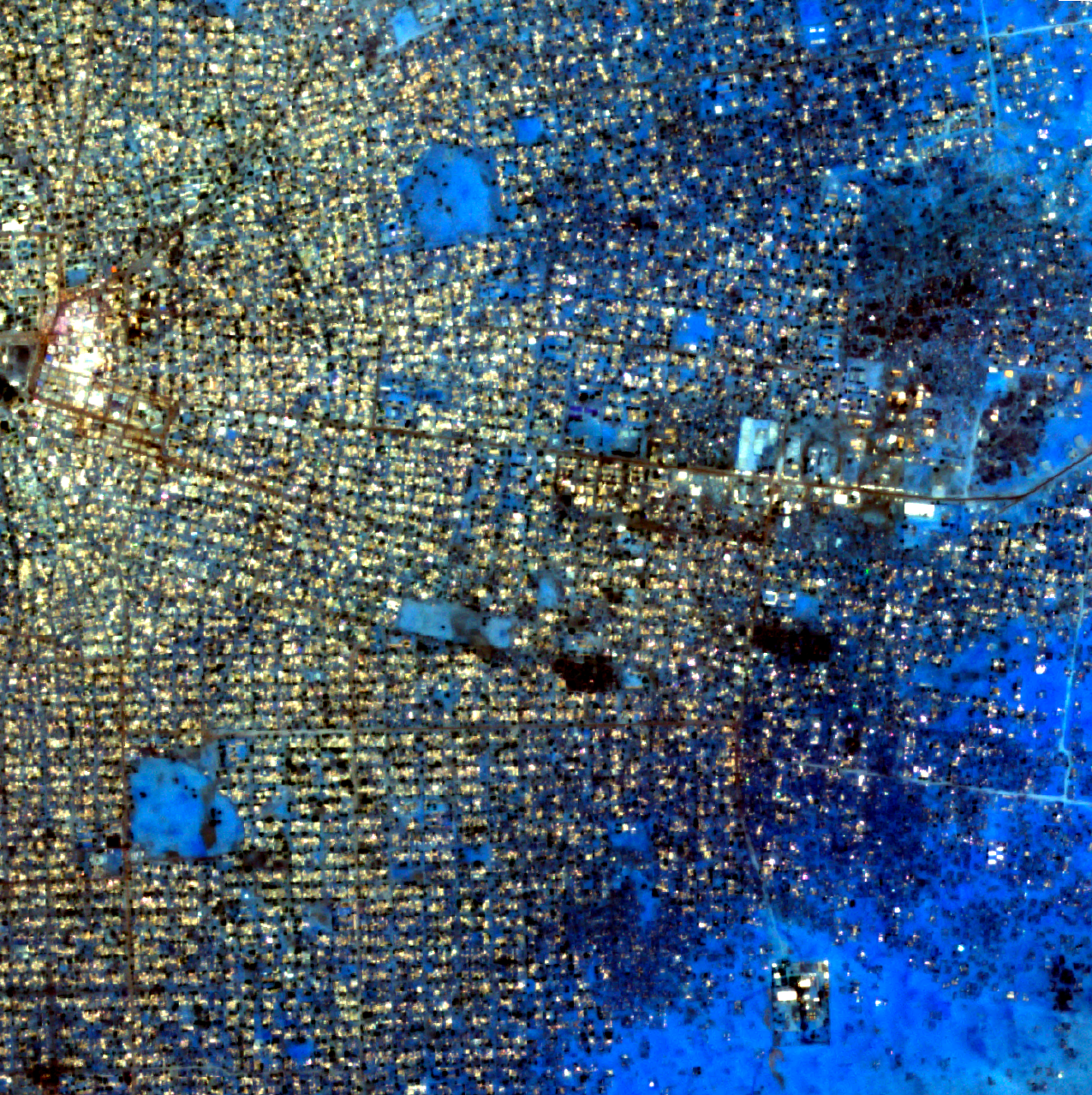}
        \caption{Before}
    \end{subfigure}
    \hfill
    \begin{subfigure}{0.32\textwidth}
        \centering
        \includegraphics[width=\linewidth]{el_fasher_after.pdf}
        \caption{After}
    \end{subfigure}
    \hfill
    \begin{subfigure}{0.32\textwidth}
        \centering
        \includegraphics[width=\linewidth]{el_fasher_pred.pdf}
        \caption{Prediction}
    \end{subfigure}
    \caption{Comparison of before and after input imagery to detect burn scars after occurrence of fire incidents in El Fasher with multiple charred places.}\label{el_fasher}
\end{figure}

\begin{figure}[h!]
    \centering
    \begin{subfigure}{0.32\textwidth}
        \centering
        \includegraphics[width=\linewidth]{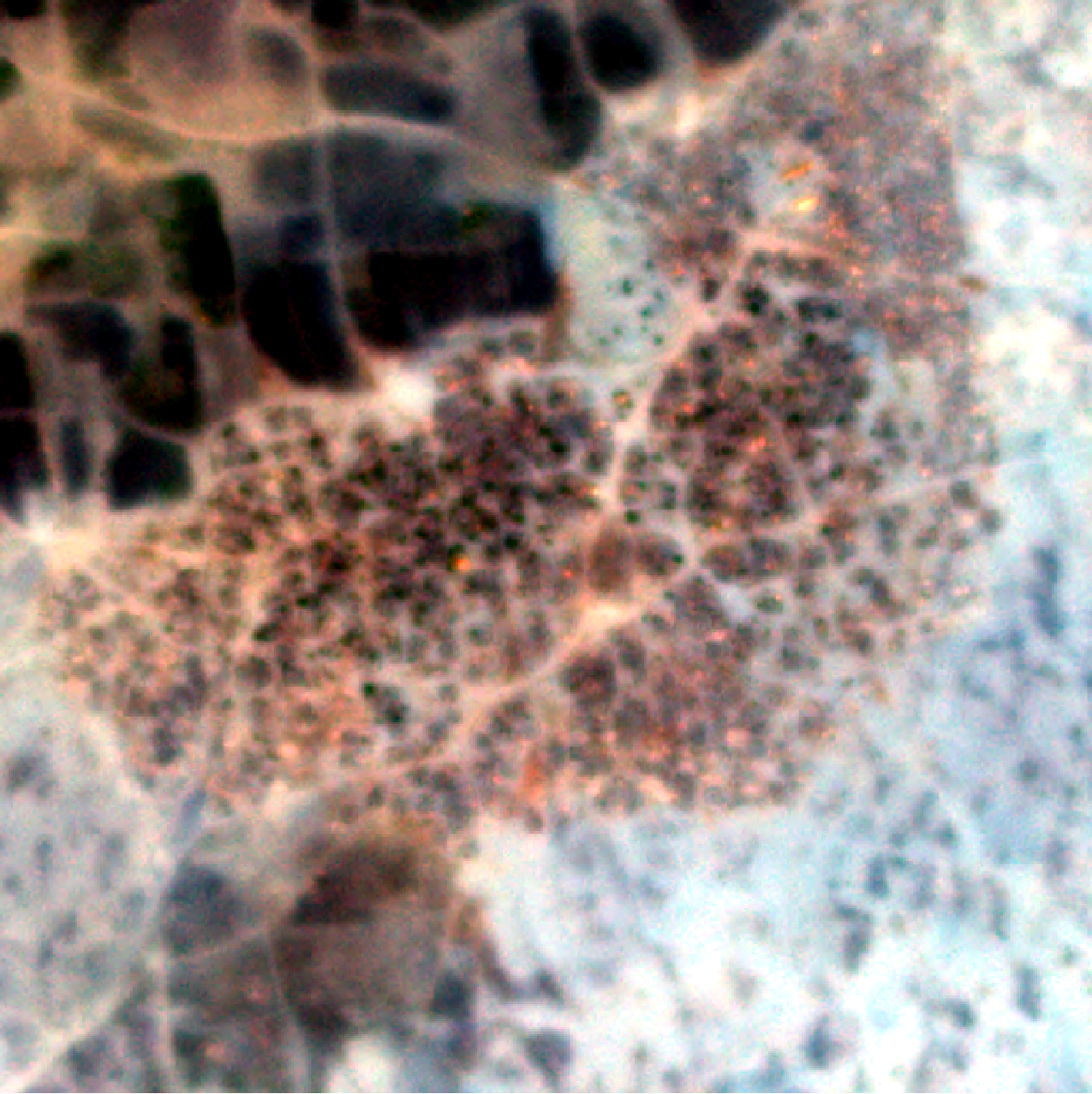}
        \caption{Before}
    \end{subfigure}
    \hfill
    \begin{subfigure}{0.32\textwidth}
        \centering
        \includegraphics[width=\linewidth]{jaranga_after.pdf}
        \caption{After}
    \end{subfigure}
    \hfill
    \begin{subfigure}{0.32\textwidth}
        \centering
        \includegraphics[width=\linewidth]{jaranga_pred.pdf}
        \caption{Prediction}
    \end{subfigure}
    \caption{Comparison of before and after input imagery with an active fire prediction in Jaranga.}\label{jaranga}
\end{figure}

\begin{figure}[h!]
    \centering
    \begin{subfigure}{0.32\textwidth}
        \centering
        \includegraphics[width=\linewidth]{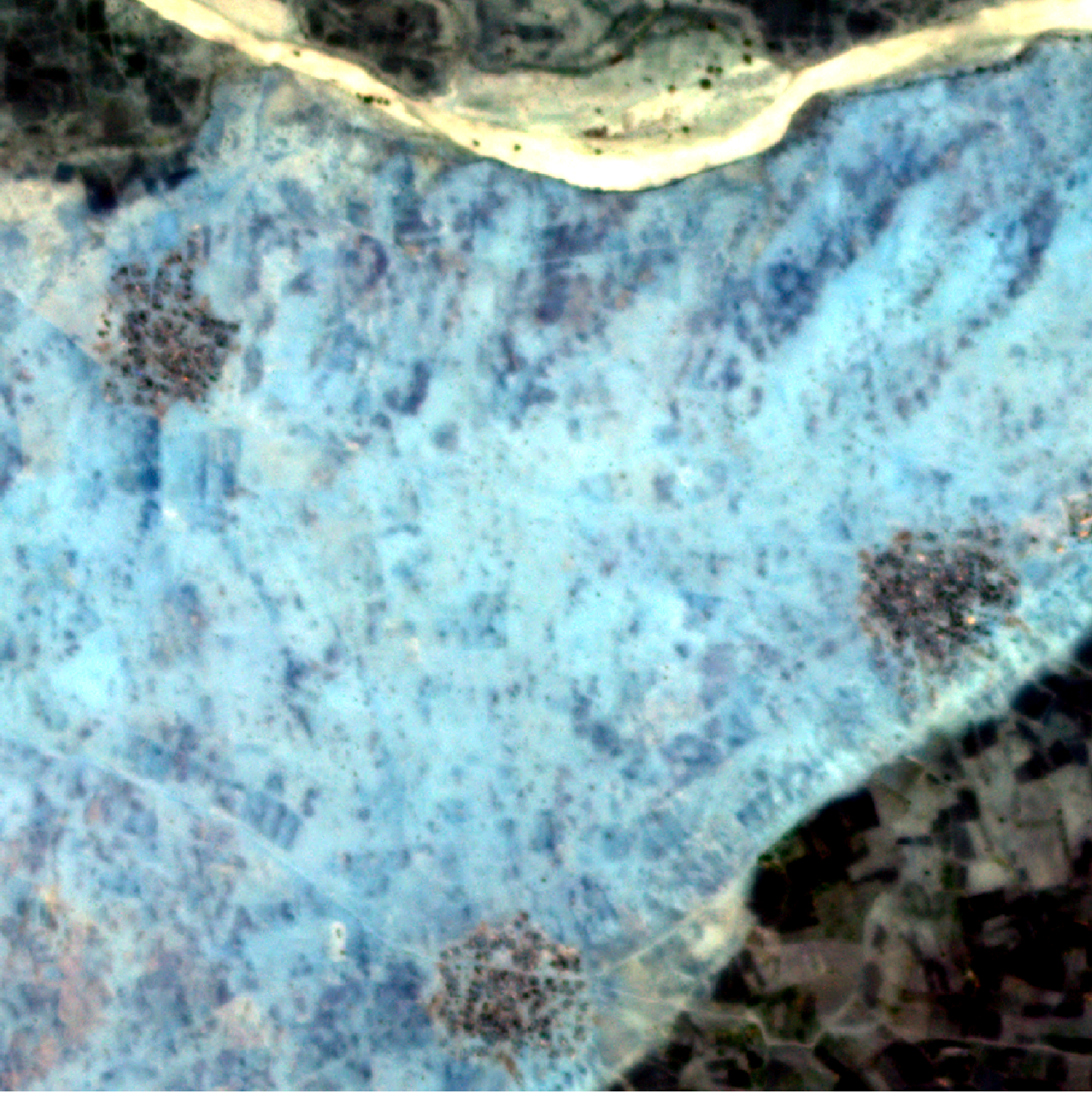}
        \caption{Before}
    \end{subfigure}
    \hfill
    \begin{subfigure}{0.32\textwidth}
        \centering
        \includegraphics[width=\linewidth]{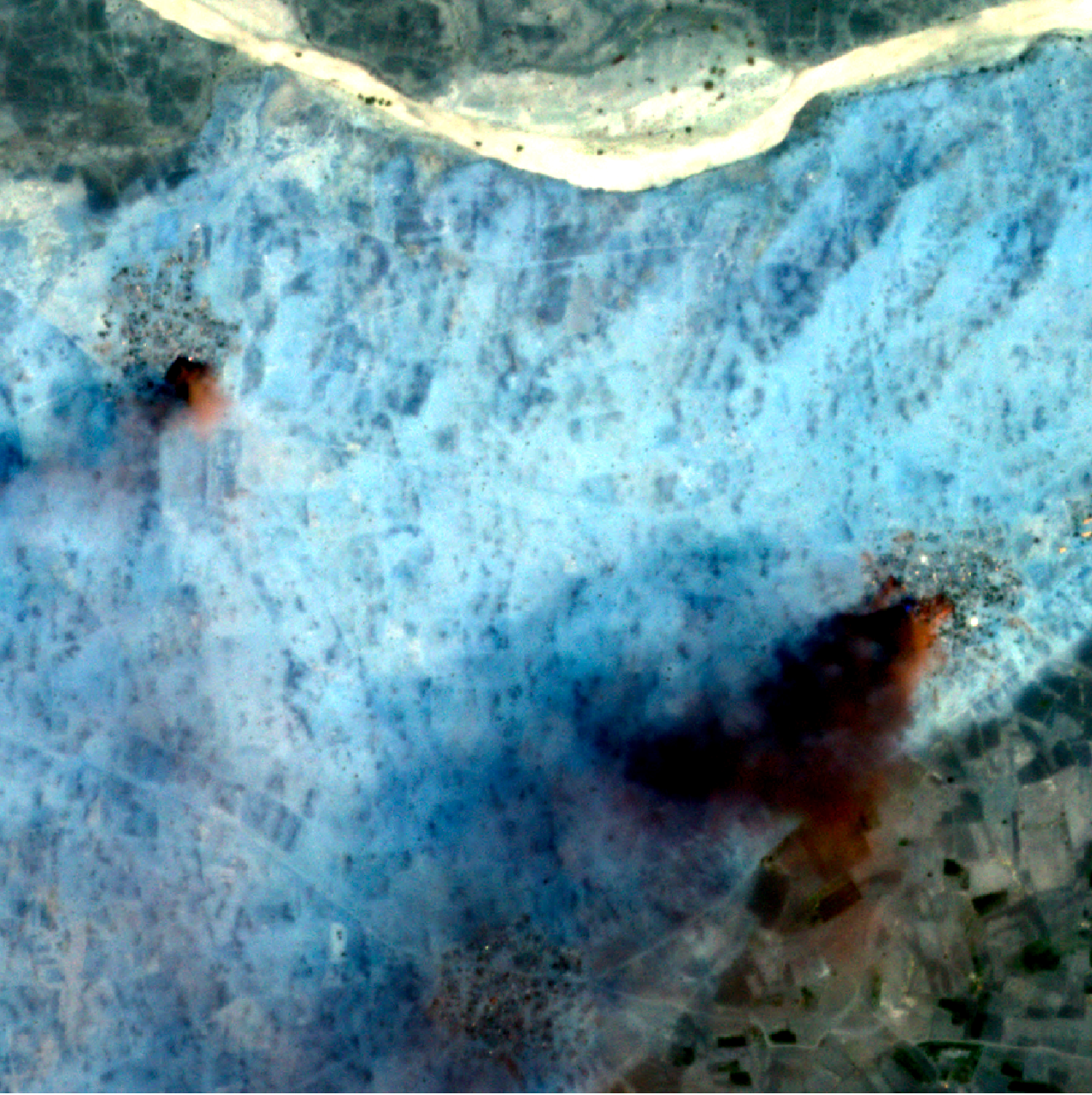}
        \caption{After}
    \end{subfigure}
    \hfill
    \begin{subfigure}{0.32\textwidth}
        \centering
        \includegraphics[width=\linewidth]{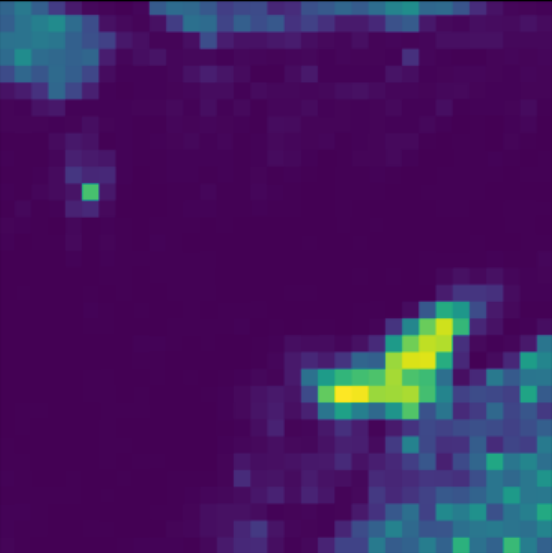}
        \caption{Prediction}
    \end{subfigure}
    \caption{Comparison of before and after input imagery with multiple closely spaced active fires in Muqrin.}\label{muqrin}
\end{figure}

\begin{figure}[h!]
    \centering
    \begin{subfigure}{0.32\textwidth}
        \centering
        \includegraphics[width=\linewidth]{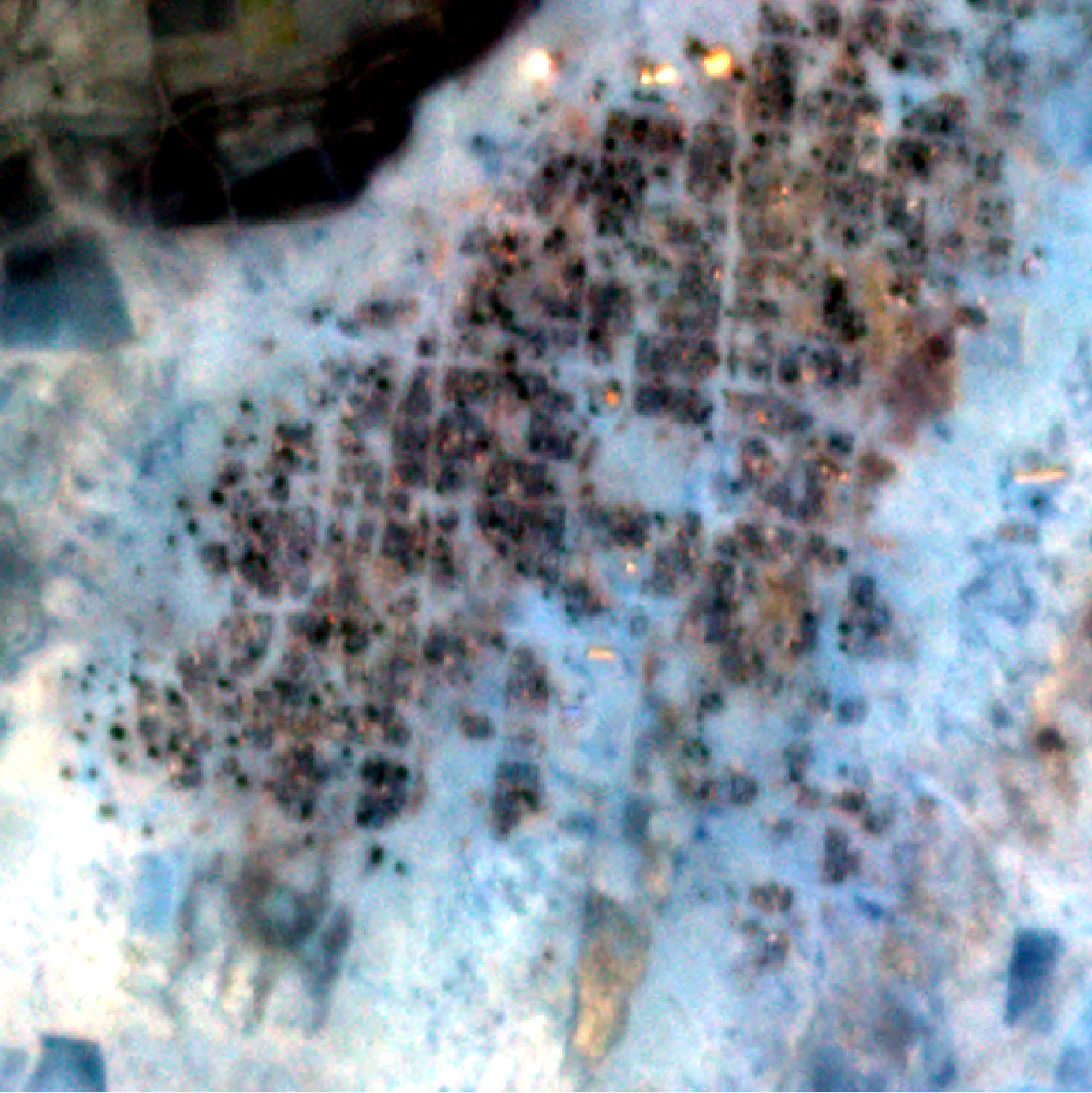}
        \caption{Before}
    \end{subfigure}
    \hfill
    \begin{subfigure}{0.32\textwidth}
        \centering
        \includegraphics[width=\linewidth]{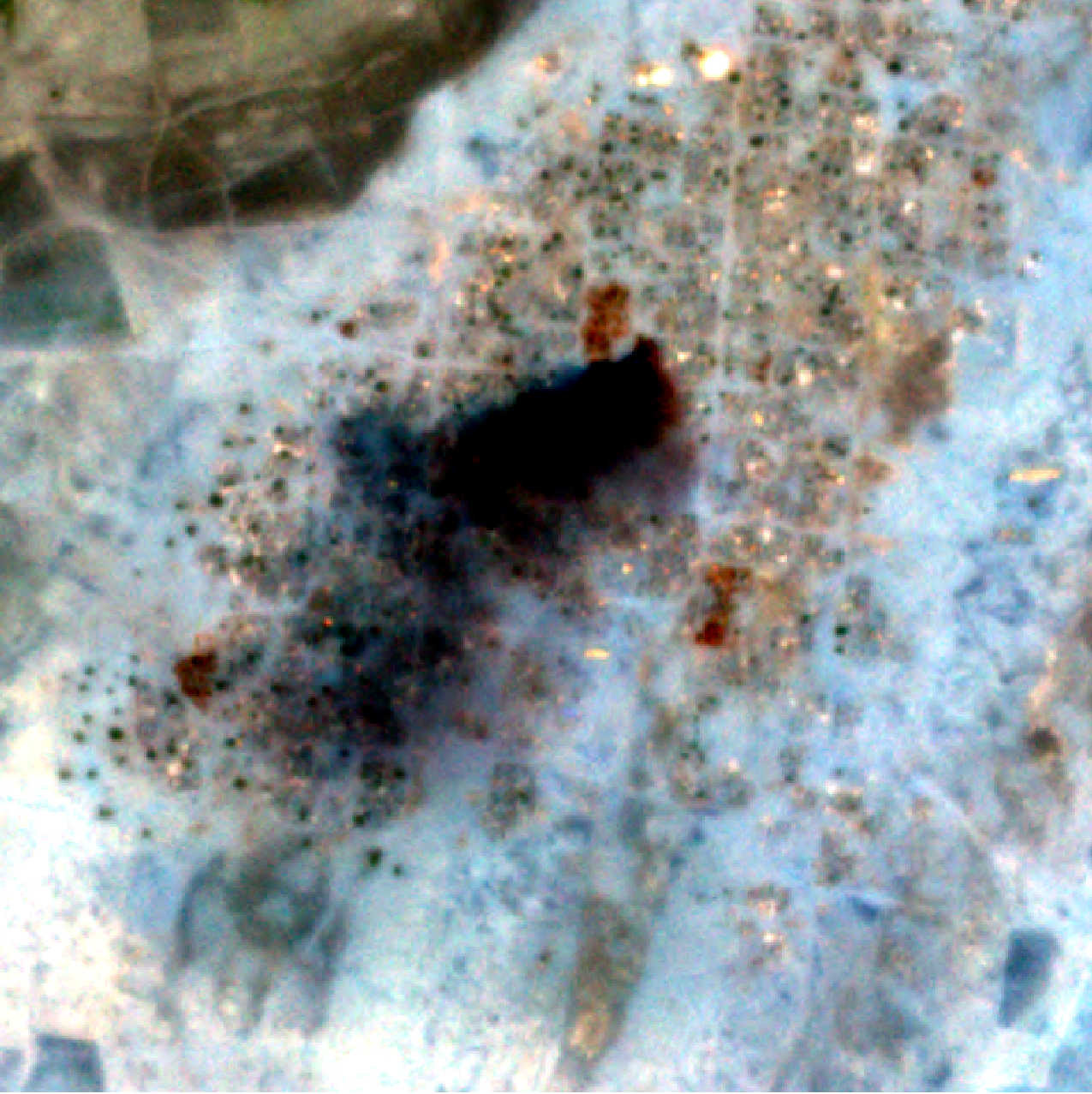}
        \caption{After}
    \end{subfigure}
    \hfill
    \begin{subfigure}{0.32\textwidth}
        \centering
        \includegraphics[width=\linewidth]{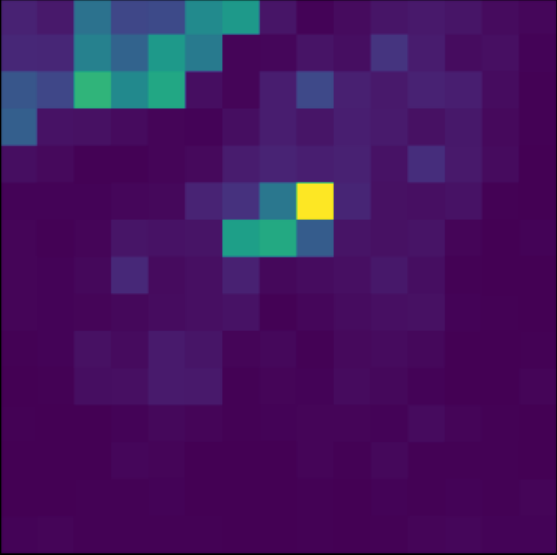}
        \caption{Prediction}
    \end{subfigure}
    \caption{Comparison of before and after input imagery with an impact of fires in Sarafaya.}\label{sarafaya}
\end{figure}

\Cref{gandahar_mkt,el_fasher,jaranga,muqrin,sarafaya} provide qualitative results across all study areas. In Gandahar Market (Figure \ref{gandahar_mkt}), the model successfully detects active fires; the input imagery also contains a visible smoke plume at the time of acquisition, which is reflected in the predicted change map. Smoke plumes are transient and depend on acquisition timing and atmospheric conditions; therefore, they are not treated as a primary or reliable indicator of fire activity in this study. Similarly, Figure \ref{el_fasher} demonstrates the model’s ability to identify burn scars following fire incidents. We note that large areas of El Fasher are burned after multiple successive fire incidents, leaving extensive swaths of charred surfaces. Even in relatively complex images, the model is generally able to delineate severely burned areas with reasonable accuracy. Figure \ref{jaranga} shows that multiple simultaneous fires are detected in Jaranga, while previously impacted locations are correctly identified in Muqrin (Figure \ref{muqrin}). Finally, Figure \ref{sarafaya} indicates accurate delineation of recently burned surfaces in Sarafaya. Across cases, predicted fire-affected regions exhibit spatial coherence and demonstrate good agreement with independently verified ground-truth labels.

\subsection{Quantitative Performance}

\begin{table}[h!]
\scriptsize
\centering
\caption{Performance metrics for Gandahar Market (median across folds).}
\label{tab:metrics_gandahar}
\begin{tabular}{llcccccc}
\toprule
\textbf{Imagery} & \textbf{Method} & \textbf{AUPRC} & \textbf{Precision} & \textbf{Recall} & \textbf{F1} & \textbf{CE} & \textbf{OE} \\
\midrule
\multirow{6}{*}{4-band} 
 & Cosine Distance & 0.56 & 0.61 & 0.53 & 0.57 & 0.39 & 0.47 \\
 & CVA & 0.58 & 0.61 & 0.52 & 0.56 & 0.39 & 0.48 \\
 & IR-MAD & 0.65 & 0.66 & 0.59 & 0.62 & 0.34 & 0.41 \\
 & dNDVI & 0.52 & 0.58 & 0.49 & 0.53 & 0.42 & 0.51 \\
 & dBAI & 0.60 & 0.63 & 0.55 & 0.59 & 0.37 & 0.45 \\
 & \textbf{LRC} & 0.74 & 0.68 & 0.75 & 0.71 & 0.32 & 0.25 \\
\midrule
\multirow{6}{*}{8-band} 
 & Cosine Distance & 0.60 & 0.64 & 0.55 & 0.59 & 0.36 & 0.45 \\
 & CVA & 0.62 & 0.64 & 0.56 & 0.60 & 0.36 & 0.44 \\
 & IR-MAD & 0.70 & 0.71 & 0.63 & 0.67 & 0.29 & 0.37 \\
 & dNDVI & 0.55 & 0.60 & 0.51 & 0.55 & 0.40 & 0.49 \\
 & dBAI & 0.66 & 0.68 & 0.60 & 0.64 & 0.32 & 0.40 \\
 & \textbf{LRC} & 0.80 & 0.72 & 0.79 & 0.75 & 0.28 & 0.21 \\
\midrule
\multirow{6}{*}{Time-series} 
 & Cosine Distance & 0.58 & 0.62 & 0.54 & 0.57 & 0.38 & 0.46 \\
 & CVA & 0.59 & 0.63 & 0.55 & 0.58 & 0.37 & 0.45 \\
 & IR-MAD & 0.66 & 0.70 & 0.62 & 0.66 & 0.30 & 0.38 \\
 & dNDVI & 0.54 & 0.59 & 0.50 & 0.54 & 0.41 & 0.50 \\
 & dBAI & 0.64 & 0.67 & 0.59 & 0.63 & 0.33 & 0.41 \\
 & \textbf{LRC} & 0.81 & 0.74 & 0.80 & 0.77 & 0.26 & 0.20 \\
\bottomrule
\end{tabular}
\end{table}

Tables \ref{tab:metrics_gandahar}--\ref{tab:metrics_sarafaya} summarize bootstrapped metrics for each location and imagery configuration, including AUPRC, precision, recall, F1-score, and commission and omission error (CE, OE) rates. The proposed VAE-based approach, referred to as latent representation change (LRC), consistently outperformed all baseline methods, including cosine distance, CVA, IR-MAD, dNDVI, and dBAI, across all sites and imagery configurations. Relative improvements in AUPRC over IR-MAD, computed as \(\frac{\mathrm{AUPRC}_{\text{LRC}} - \mathrm{AUPRC}_{\text{IR-MAD}}}{\mathrm{AUPRC}_{\text{IR-MAD}}}\), ranged from 14\% to 36\%, with sites having clearer fire signatures (e.g., Muqrin and Sarafaya) achieving AUPRC values exceeding 0.80.  Although these case studies correspond to confirmed conflict events, fire-affected tiles comprise less than 10\% of the scene area. The remaining tiles establish the nominal background distribution, representing unburned urban surfaces, vegetation, and other background variability.

\begin{table}[h!]
\scriptsize
\centering
\caption{Performance metrics for El Fasher.}
\label{tab:metrics_elfasher}
\begin{tabular}{llcccccc}
\toprule
\textbf{Imagery} & \textbf{Method} & \textbf{AUPRC} & \textbf{Precision} & \textbf{Recall} & \textbf{F1} & \textbf{CE} & \textbf{OE} \\
\midrule
\multirow{6}{*}{4-band} 
 & Cosine Distance & 0.43 & 0.51 & 0.45 & 0.48 & 0.49 & 0.55 \\
 & CVA & 0.48 & 0.51 & 0.44 & 0.47 & 0.49 & 0.56 \\
 & IR-MAD & 0.50 & 0.56 & 0.50 & 0.53 & 0.44 & 0.50 \\
 & dNDVI & 0.39 & 0.47 & 0.41 & 0.44 & 0.53 & 0.59 \\
 & dBAI & 0.46 & 0.52 & 0.47 & 0.49 & 0.48 & 0.53 \\
 & \textbf{LRC} & 0.68 & 0.62 & 0.66 & 0.64 & 0.38 & 0.34 \\
\midrule
\multirow{6}{*}{8-band} 
 & Cosine Distance & 0.44 & 0.48 & 0.42 & 0.45 & 0.52 & 0.58 \\
 & CVA & 0.46 & 0.49 & 0.43 & 0.46 & 0.51 & 0.57 \\
 & IR-MAD & 0.49 & 0.54 & 0.48 & 0.51 & 0.46 & 0.52 \\
 & dNDVI & 0.41 & 0.48 & 0.42 & 0.45 & 0.52 & 0.58 \\
 & dBAI & 0.50 & 0.55 & 0.49 & 0.52 & 0.45 & 0.51 \\
 & \textbf{LRC} & 0.62 & 0.57 & 0.63 & 0.60 & 0.43 & 0.37 \\
\midrule
\multirow{6}{*}{Time-series} 
 & Cosine Distance & 0.40 & 0.48 & 0.42 & 0.45 & 0.52 & 0.58 \\
 & CVA & 0.41 & 0.49 & 0.43 & 0.46 & 0.51 & 0.57 \\
 & IR-MAD & 0.48 & 0.55 & 0.50 & 0.52 & 0.45 & 0.50 \\
 & dNDVI & 0.40 & 0.47 & 0.41 & 0.44 & 0.53 & 0.59 \\
 & dBAI & 0.48 & 0.53 & 0.48 & 0.50 & 0.47 & 0.52 \\
 & \textbf{LRC} & 0.65 & 0.60 & 0.63 & 0.61 & 0.40 & 0.37 \\
\bottomrule
\end{tabular}
\end{table}

In the heterogeneous urban environment of El Fasher, the LRC framework achieved a 36\% relative improvement of AUPRC over IR-MAD, indicating improved discrimination of fire-affected regions in complex urban scenes. The model achieved a recall of 0.66 for independently verified fire-affected tiles, with omission errors primarily associated with small or weak burn signatures that were difficult to distinguish at the 3 m spatial resolution, as well as spectral ambiguity arising from successive fire incidents. A precision of 0.62 reflects false positives associated with spectrally similar urban materials such as dark rooftops and shadows, which are particularly challenging to distinguish in 4-band imagery. Despite these challenges, the resulting F1-score of 0.64 demonstrates an improvement over pixel-based baselines such as dNDVI (0.44), highlighting the benefit of learned latent representations in complex urban settings.

The reported precision and recall values in Table~\ref{tab:metrics_elfasher} and other case studies should be interpreted as measures of agreement with independently verified fire-affected labels rather than indicators of fire-specific physical attribution. In this context, precision quantifies the proportion of detected anomalies aligning with verified fire-affected regions, while recall quantifies the sensitivity to such regions. The moderate correlation with dBAI (Table~\ref{tab:cosine_dBAI_correlation}) further suggests that the model captures gradients of burn-related change, while remaining influenced by other sources of spectral–spatial variability.

\begin{table}[b!]
\scriptsize
\centering
\caption{Performance metrics for Muqrin.}
\label{tab:metrics_muqrin}
\begin{tabular}{llcccccc}
\toprule
\textbf{Imagery} & \textbf{Method} & \textbf{AUPRC} & \textbf{Precision} & \textbf{Recall} & \textbf{F1} & \textbf{CE} & \textbf{OE} \\
\midrule
\multirow{6}{*}{4-band} 
 & Cosine Distance & 0.62 & 0.66 & 0.59 & 0.62 & 0.34 & 0.41 \\
 & CVA & 0.63 & 0.67 & 0.58 & 0.62 & 0.33 & 0.42 \\
 & IR-MAD & 0.65 & 0.64 & 0.70 & 0.67 & 0.36 & 0.30 \\
 & dNDVI & 0.58 & 0.63 & 0.55 & 0.59 & 0.37 & 0.45 \\
 & dBAI & 0.64 & 0.66 & 0.68 & 0.67 & 0.34 & 0.32 \\
 & \textbf{LRC} & 0.77 & 0.79 & 0.82 & 0.80 & 0.21 & 0.18 \\
\midrule
\multirow{6}{*}{8-band} 
 & Cosine Distance & 0.65 & 0.68 & 0.62 & 0.66 & 0.32 & 0.38 \\
 & CVA & 0.67 & 0.69 & 0.61 & 0.65 & 0.31 & 0.39 \\
 & IR-MAD & 0.69 & 0.67 & 0.72 & 0.69 & 0.33 & 0.28 \\
 & dNDVI & 0.61 & 0.66 & 0.58 & 0.62 & 0.34 & 0.42 \\
 & dBAI & 0.68 & 0.69 & 0.71 & 0.70 & 0.31 & 0.29 \\
 & \textbf{LRC} & 0.79 & 0.81 & 0.83 & 0.82 & 0.19 & 0.17 \\
\midrule
\multirow{6}{*}{Time-series} 
 & Cosine Distance & 0.64 & 0.69 & 0.62 & 0.65 & 0.31 & 0.38 \\
 & CVA & 0.65 & 0.70 & 0.63 & 0.66 & 0.30 & 0.37 \\
 & IR-MAD & 0.71 & 0.69 & 0.74 & 0.71 & 0.31 & 0.26 \\
 & dNDVI & 0.60 & 0.65 & 0.57 & 0.61 & 0.35 & 0.43 \\
 & dBAI & 0.69 & 0.70 & 0.72 & 0.71 & 0.30 & 0.28 \\
 & \textbf{LRC} & 0.81 & 0.83 & 0.85 & 0.83 & 0.17 & 0.15 \\
\bottomrule
\end{tabular}
\end{table}

Among the baseline methods, IR-MAD generally achieved the best performance, effectively modeling multivariate change while suppressing invariant background structure, particularly in Gandahar Market, Muqrin, and Sarafaya. CVA also contributed to robust detection in these sites by capturing multivariate change, though with slightly lower accuracy. The dBAI index performed competitively in scenes with clear burn signatures, sometimes approaching IR-MAD, while dNDVI consistently underperformed, particularly in urban or sparsely vegetated areas such as El Fasher and Jaranga, reflecting its reliance on vegetation loss as a proxy for fire damage. Cosine distance and CVA exhibited moderate performance overall but were more sensitive to radiometric variability and scene heterogeneity.

The inclusion of 8-band and time-series imagery modestly improved both dNDVI and dBAI, particularly dBAI, by enhancing spectral differentiation between burned and unburned surfaces. Although these indices do not fully utilize all 8 bands, they benefit from the additional red-edge, coastal, blue, and yellow bands. Red-edge bands enhance sensitivity to subtle vegetation stress and structural changes, helping distinguish fire-affected vegetation from unburned areas. Coastal, blue, and yellow bands improve atmospheric and surface characterization, crucial in complex environments where soil and residual vegetation may cause confusion. However, since these indices are pixel-based and lack spatial and temporal context, their ability to detect nuanced fire damage over time remains limited, particularly in heterogeneous landscapes with variable fire impacts.

\begin{table}[t!]
\scriptsize
\centering
\caption{Performance metrics for Jaranga.}
\label{tab:metrics_jaranga}
\begin{tabular}{llcccccc}
\toprule
\textbf{Imagery} & \textbf{Method} & \textbf{AUPRC} & \textbf{Precision} & \textbf{Recall} & \textbf{F1} & \textbf{CE} & \textbf{OE} \\
\midrule
\multirow{6}{*}{4-band} 
 & Cosine Distance & 0.41 & 0.49 & 0.42 & 0.45 & 0.51 & 0.58 \\
 & CVA & 0.46 & 0.49 & 0.41 & 0.45 & 0.51 & 0.59 \\
 & IR-MAD & 0.52 & 0.54 & 0.48 & 0.51 & 0.46 & 0.52 \\
 & dNDVI & 0.38 & 0.46 & 0.40 & 0.43 & 0.54 & 0.60 \\
 & dBAI & 0.49 & 0.53 & 0.47 & 0.50 & 0.47 & 0.53 \\
 & \textbf{LRC} & 0.65 & 0.60 & 0.64 & 0.62 & 0.40 & 0.36 \\
\midrule
\multirow{6}{*}{8-band} 
 & Cosine Distance & 0.49 & 0.55 & 0.48 & 0.51 & 0.45 & 0.52 \\
 & CVA & 0.54 & 0.55 & 0.48 & 0.51 & 0.45 & 0.52 \\
 & IR-MAD & 0.63 & 0.63 & 0.57 & 0.60 & 0.37 & 0.43 \\
 & dNDVI & 0.44 & 0.50 & 0.43 & 0.46 & 0.50 & 0.57 \\
 & dBAI & 0.60 & 0.62 & 0.56 & 0.59 & 0.38 & 0.44 \\
 & \textbf{LRC} & 0.72 & 0.67 & 0.69 & 0.68 & 0.33 & 0.31 \\
\midrule
\multirow{6}{*}{Time-series} 
 & Cosine Distance & 0.46 & 0.52 & 0.45 & 0.48 & 0.48 & 0.55 \\
 & CVA & 0.47 & 0.53 & 0.46 & 0.49 & 0.47 & 0.54 \\
 & IR-MAD & 0.55 & 0.60 & 0.54 & 0.57 & 0.40 & 0.46 \\
 & dNDVI & 0.42 & 0.49 & 0.42 & 0.45 & 0.51 & 0.58 \\
 & dBAI & 0.56 & 0.60 & 0.54 & 0.57 & 0.40 & 0.46 \\
 & \textbf{LRC} & 0.74 & 0.68 & 0.71 & 0.69 & 0.32 & 0.29 \\
\bottomrule
\end{tabular}
\end{table}

Across all sites, metrics, and imagery types, the 95\% confidence intervals were generally narrow, with a median span of 0.01 to 0.04, reflecting low to moderate variability in the bootstrapped estimates for most scenes, although resampling within scenes may underestimate uncertainty if observations are not fully independent. In the highly heterogeneous urban setting of El Fasher, wider intervals were observed, consistent with its more complex scene structure.

\begin{table}[b!]
\scriptsize
\centering
\caption{Performance metrics for Sarafaya.}
\label{tab:metrics_sarafaya}
\begin{tabular}{llcccccc}
\toprule
\textbf{Imagery} & \textbf{Method} & \textbf{AUPRC} & \textbf{Precision} & \textbf{Recall} & \textbf{F1} & \textbf{CE} & \textbf{OE} \\
\midrule
\multirow{6}{*}{4-band} 
 & Cosine Distance & 0.64 & 0.69 & 0.62 & 0.65 & 0.31 & 0.38 \\
 & CVA & 0.67 & 0.70 & 0.61 & 0.66 & 0.30 & 0.39 \\
 & IR-MAD & 0.68 & 0.72 & 0.66 & 0.69 & 0.28 & 0.34 \\
 & dNDVI & 0.60 & 0.65 & 0.57 & 0.61 & 0.35 & 0.43 \\
 & dBAI & 0.67 & 0.71 & 0.65 & 0.68 & 0.29 & 0.35 \\
 & \textbf{LRC} & 0.78 & 0.80 & 0.83 & 0.81 & 0.20 & 0.17 \\
\midrule
\multirow{6}{*}{8-band} 
 & Cosine Distance & 0.64 & 0.68 & 0.63 & 0.65 & 0.32 & 0.37 \\
 & CVA & 0.69 & 0.71 & 0.63 & 0.67 & 0.29 & 0.37 \\
 & IR-MAD & 0.70 & 0.75 & 0.69 & 0.72 & 0.25 & 0.31 \\
 & dNDVI & 0.63 & 0.68 & 0.60 & 0.64 & 0.32 & 0.40 \\
 & dBAI & 0.69 & 0.73 & 0.68 & 0.70 & 0.27 & 0.32 \\
 & \textbf{LRC} & 0.80 & 0.81 & 0.84 & 0.83 & 0.19 & 0.16 \\
\midrule
\multirow{6}{*}{Time-series} 
 & Cosine Distance & 0.65 & 0.71 & 0.64 & 0.67 & 0.29 & 0.36 \\
 & CVA & 0.66 & 0.72 & 0.65 & 0.68 & 0.28 & 0.35 \\
 & IR-MAD & 0.70 & 0.77 & 0.72 & 0.74 & 0.23 & 0.28 \\
 & dNDVI & 0.62 & 0.67 & 0.60 & 0.63 & 0.33 & 0.40 \\
 & dBAI & 0.70 & 0.75 & 0.71 & 0.73 & 0.25 & 0.29 \\
 & \textbf{LRC} & 0.82 & 0.84 & 0.86 & 0.84 & 0.16 & 0.14 \\
\bottomrule
\end{tabular}
\end{table}

\begin{table}[h!]
\scriptsize
\centering
\caption{Cohen’s $d$ effect sizes for LRC against IR-MAD across all metrics.}
\label{tab:metrics_cohend}
\begin{tabular}{llcccc}
\toprule
\multirow{2}{*}{} \textbf{Place} & \textbf{Imagery} & \textbf{Cohen's $d$} & \textbf{Cohen's $d$} & \textbf{Cohen's $d$} & \textbf{Cohen's $d$} \\
 & & \textbf{(AUPRC)} & \textbf{(Precision)} & \textbf{(Recall)} & \textbf{(F1)} \\
\midrule
\multirow{3}{*}{\parbox{1cm}{\centering Gandahar\\Market}}
 & 4-band      & 1.15 & 0.45 & 1.30 & 1.10 \\
 & 8-band      & 1.18 & 0.48 & 1.32 & 1.12 \\
 & Time-series & 1.16 & 0.46 & 1.31 & 1.11 \\
\midrule
\multirow{3}{*}{El Fasher} & 4-band & 1.10 & 0.48 & 1.25 & 1.05 \\
& 8-band & 0.75 & 0.35 & 0.85 & 0.70 \\
& Time-series & 1.05 & 0.45 & 1.15 & 0.98 \\
\midrule
\multirow{3}{*}{Muqrin} & 4-band & 1.25 & 0.50 & 1.35 & 1.20 \\
& 8-band & 1.27 & 0.52 & 1.36 & 1.22 \\
& Time-series & 1.26 & 0.51 & 1.35 & 1.21 \\
\midrule
\multirow{3}{*}{Jaranga} & 4-band & 1.05 & 0.42 & 1.20 & 1.00 \\
& 8-band & 1.08 & 0.45 & 1.22 & 1.02 \\
& Time-series & 1.06 & 0.44 & 1.21 & 1.01 \\
\midrule
\multirow{3}{*}{Sarafaya} & 4-band & 1.20 & 0.48 & 1.30 & 1.15 \\
& 8-band & 1.22 & 0.50 & 1.31 & 1.16 \\
& Time-series & 1.21 & 0.49 & 1.30 & 1.15 \\
\bottomrule
\end{tabular}
\end{table}

In addition to absolute performance metrics, in Table \ref{tab:metrics_cohend} we quantified the magnitude of improvement of the VAE-based approach over IR-MAD using Cohen’s $d$, since other models generally exhibited lower performance. Following the paired tile-level experimental design, effect sizes were computed using the standard deviation of paired differences across corresponding $32 \times 32$ tiles within each bootstrap sample. Effect sizes ranged from 0.35 to 1.36 across metrics, with the largest values observed for Recall ($d \approx 1.30$), reflecting the framework’s design to prioritize detection of fire-affected areas while maintaining viable precision. Paired Wilcoxon signed-rank tests indicated that these differences were statistically significant ($p < 0.01$), consistent with the magnitude of the estimated effect sizes.

To validate the robustness of the proposed approach with respect to similarity metric selection, we conduct an ablation study on latent-space distance measures. Table \ref{tab:ablation_distance} evaluates the sensitivity of latent representation change (LRC) to the choice of distance metric, noting that the similarity measure defines the geometry of comparisons in latent space. We compare cosine distance, Euclidean distance, Mahalanobis distance, and Jensen--Shannon (JS) divergence, the latter computed between Gaussian distributions parameterized by $(\mu,\sigma)$ from the VAE.

\begin{table}[h!]
\scriptsize
\centering
\caption{Ablation study of distance metrics for latent representation change (LRC) detection using 4-band imagery. All methods use identical latent embeddings and preprocessing, with a 95th percentile threshold applied for binary classification metrics.}
\label{tab:ablation_distance}
\begin{tabular}{llcccccc}
\toprule
\textbf{Place} & \textbf{Metric} & \textbf{AUPRC} & \textbf{Precision} & \textbf{Recall} & \textbf{F1} & \textbf{CE} & \textbf{OE} \\
\midrule

\multirow{4}{*}{Gandahar Market}
 & Cosine        & 0.74 & 0.68 & 0.75 & 0.71 & 0.32 & 0.25 \\
 & Euclidean     & 0.70 & 0.66 & 0.71 & 0.68 & 0.34 & 0.29 \\
 & Mahalanobis   & 0.73 & 0.69 & 0.73 & 0.71 & 0.31 & 0.27 \\
 & JS Divergence & 0.67 & 0.63 & 0.69 & 0.66 & 0.37 & 0.31 \\

\midrule

\multirow{4}{*}{El Fasher}
 & Cosine        & 0.68 & 0.62 & 0.66 & 0.64 & 0.38 & 0.34 \\
 & Euclidean     & 0.64 & 0.60 & 0.62 & 0.61 & 0.40 & 0.38 \\
 & Mahalanobis   & 0.67 & 0.61 & 0.65 & 0.63 & 0.39 & 0.35 \\
 & JS Divergence & 0.61 & 0.57 & 0.60 & 0.58 & 0.43 & 0.40 \\

\midrule

\multirow{4}{*}{Muqrin}
 & Cosine        & 0.77 & 0.79 & 0.82 & 0.80 & 0.21 & 0.18 \\
 & Euclidean     & 0.73 & 0.77 & 0.78 & 0.77 & 0.23 & 0.22 \\
 & Mahalanobis   & 0.76 & 0.80 & 0.80 & 0.80 & 0.20 & 0.20 \\
 & JS Divergence & 0.70 & 0.74 & 0.75 & 0.74 & 0.26 & 0.25 \\

\midrule

\multirow{4}{*}{Jaranga}
 & Cosine        & 0.65 & 0.60 & 0.64 & 0.62 & 0.40 & 0.36 \\
 & Euclidean     & 0.61 & 0.58 & 0.60 & 0.59 & 0.42 & 0.40 \\
 & Mahalanobis   & 0.64 & 0.61 & 0.63 & 0.62 & 0.39 & 0.37 \\
 & JS Divergence & 0.58 & 0.55 & 0.58 & 0.56 & 0.45 & 0.42 \\

\midrule

\multirow{4}{*}{Sarafaya}
 & Cosine        & 0.78 & 0.80 & 0.83 & 0.81 & 0.20 & 0.17 \\
 & Euclidean     & 0.74 & 0.78 & 0.79 & 0.78 & 0.22 & 0.21 \\
 & Mahalanobis   & 0.77 & 0.81 & 0.82 & 0.81 & 0.19 & 0.18 \\
 & JS Divergence & 0.71 & 0.75 & 0.77 & 0.76 & 0.25 & 0.23 \\

\bottomrule
\end{tabular}
\end{table}

Across all case studies, cosine distance yields consistent performance, achieving near-highest AUPRC while maintaining balanced precision and recall. This behavior is expected, as cosine distance is invariant to vector magnitude and emphasizes angular deviations, making it robust to residual radiometric variability and normalization effects. In contrast, Euclidean distance is sensitive to feature magnitude and consistently underperforms, while JS divergence provides the lowest accuracy, indicating limited benefit from distribution-based comparisons in the latent space. Mahalanobis distance yields comparable performance in several cases, occasionally matching cosine in F1-score and AUPRC. However, this similarity comes at the cost of increased computational overhead and sensitivity to covariance estimation, which can be unstable in small or heterogeneous samples. 

Importantly, performance differences across metrics are modest (typically within 3--5\% AUPRC), indicating that improvements are primarily driven by the learned latent representations rather than the specific distance metric. Within this context, cosine distance provides a simple, stable, and computationally efficient choice, achieving performance comparable to Mahalanobis without requiring covariance estimation. Additionally, prior work has demonstrated the effectiveness of cosine similarity as a robust measure in normalized embedding spaces \cite{zhang2022beyond, wojke2018deep}.

To assess whether latent-space anomaly scores reflect the magnitude of fire impact, we compute Spearman rank correlations between tile-level cosine distance scores and the differenced Burn Area Index (dBAI), used as a proxy for burn severity \cite{claverie2015evaluation}. As shown in Table~\ref{tab:cosine_dBAI_correlation}, the correlations between tile-level cosine distance scores and dBAI values are moderate across all case studies ($\rho = 0.52$--$0.68$), reflecting real-world variability at the tile scale. Lower correlations in heterogeneous urban scenes (e.g., El Fasher) are attributable to mixed pixels, successive incidents, and spectral noise, whereas more homogeneous burn areas (e.g., Muqrin and Sarafaya) exhibit stronger alignment.

\begin{table}[h!]
\scriptsize
\centering
\caption{Spearman rank correlation ($\rho$) between tile-level cosine distance and dBAI values for 4-band imagery across the five case studies.}
\label{tab:cosine_dBAI_correlation}
\begin{tabular}{lc}
\toprule
\textbf{Case Study} & \textbf{Spearman $\rho$} \\
\midrule
Gandahar Market & 0.63 \\
El Fasher       & 0.52 \\
Muqrin         & 0.68 \\
Jaranga        & 0.60 \\
Sarafaya       & 0.66 \\
\bottomrule
\end{tabular}
\end{table}

While dBAI provides a physically interpretable proxy for burn intensity, it is not a direct measure of fire severity and may be influenced by background materials. Within this context, the observed correlations indicate that cosine distance captures meaningful gradients of fire-related change, with larger latent-space deviations generally associated with stronger burn signatures. Accordingly, cosine distance should be interpreted as a relative indicator of change magnitude rather than a direct proxy for fire severity.

\subsection{Observations on False Positives and False Negatives}
Analysis of tile-level outputs revealed systematic patterns in prediction errors. False positives were primarily linked to residual burned material from prior incidents or debris resembling charred surfaces, reflecting the model's sensitivity to broader damage signatures. Instances where active fires or burn scars went undetected were primarily restricted to small, fragmented events physically smaller than the 3 m spatial resolution of the PlanetScope sensor. These omissions are an inherent hardware limitation of optical remote sensing in densely built urban areas. These observational constraints align with known limitations of optical imagery in conflict environments, and are considered an acceptable trade-off in operational settings that prioritize rapid, broad-scale situational awareness over exhaustive ground-truth error minimization \cite{zwijnenburg2023leveraging}. Despite these sensor-level challenges, overall model recall remained high, supporting the framework’s near–real-time monitoring objective of sensitive detection of fire-affected areas.

More generally, the proposed framework detects anomalous spectral–spatial change and does not explicitly distinguish between underlying physical causes. As a result, non-fire processes that produce strong surface changes, such as rapid vegetation loss (e.g., harvesting), urban construction, soil exposure, and hydrological variability due to rainfall-driven surface water or flooding, may lead to false positives. These phenomena can exhibit spectral responses partially overlapping with burn signatures in optical imagery. However, they often differ in spatial structure and context: hydrological changes tend to be spatially diffuse or constrained by topography \cite{homer2020conterminous}, while agricultural or construction-related changes are typically more regular or temporally gradual \cite{yin2018mapping}. In contrast, fire damage more commonly produces irregular, high-contrast burn scars that are spatially coherent and aligned with known incident locations, which helps mitigate such ambiguities \cite{ch2004landscape, liu2014study}.

\subsection{Sufficiency of 4-band Imagery}
Across all locations, 4-band PlanetScope imagery captured sufficient spectral information for fire detection. While modest improvements were observed with 8-band imagery or short temporal sequences (median relative AUPRC gains of 3--7\% across tested scenes), paired statistical tests indicated that these gains were small and often not operationally significant. We note that this observation is specific to the scenes and fire types analyzed; additional spectral bands may provide greater benefits in regions with more subtle fire signatures or heterogeneous land cover. In El Fasher, both the 8-band and time-series configurations yielded slightly lower AUPRC and Recall than the 4-band imagery, likely due to increased sensitivity to inter-band noise and residual misregistration errors in highly heterogeneous urban settings.

\subsection{Summary}

Overall, the proposed VAE-based approach indicates consistent and statistically significant improvements over CVA, cosine distance, and IR-MAD. Bootstrapped confidence intervals support stability across scenes, effect sizes suggest meaningful performance gains, and observed error patterns provide actionable insight for near–real-time monitoring of conflict-related fires. All experiments are compatible with the near–real-time processing pipeline described in Section \ref{methodology_eval}, producing outputs within 24–30 hours of image acquisition. Consistent with prior work \cite{ruuvzivcka2022ravaen, dong2020self}, these results indicate that latent representation–based change detection tends to improve performance over conventional pixel-difference and transformation-based methods. Considering the acquisition, preprocessing, and computational overheads of multi-spectral and multi-temporal inputs, the limited performance gains observed with additional data modalities suggest that the proposed lightweight 4-band configuration offers a favorable trade-off between accuracy and efficiency for near–real-time conflict monitoring. We note that all quantitative results are conditioned on the availability of cloud-free optical imagery and accurate co-registration between pre- and post-incident scenes.

\section{Discussion}

Our findings demonstrate that near–real-time detection of conflict-related fire damage is feasible using near-daily PlanetScope imagery combined with latent representation–based change detection. Across multiple case studies, the proposed VAE-based approach yields improved performance relative to CVA, cosine distance, dNDVI, dBAI, and IR-MAD, producing spatially coherent outputs suitable for rapid situational awareness. These findings align with prior research demonstrating the potential of latent representation learning in satellite-based change detection \cite{bennett2022improving}. Additionally, results indicate that 4-band imagery captures sufficient discriminative information required for fire detection, with additional spectral bands or short temporal sequences providing only modest gains.

Furthermore, training on the ecologically distinct WorldFloods dataset reduces the risk of domain-specific overfitting, as the unsupervised VAE has no prior exposure to Sudan’s fire patterns or urban morphology. The system is intended to verify reported incidents, so evaluation on geolocated reports aligns with operational use, where focusing on known locations is an acceptable trade-off for timely validation in data-scarce regions.

Compared with public satellite platforms such as MODIS, VIIRS, and Sentinel-2, PlanetScope provides a practical balance between spatial resolution and revisit frequency that is particularly advantageous in rapidly evolving conflict environments. While Sentinel-2 offers 13 spectral bands, including SWIR channels commonly used for burn detection, its effective revisit interval (often 5–10 days) can limit timely assessment of rapidly changing fire events. PlanetScope’s near-daily 3 m imagery reduces temporal gaps and facilitates more continuous monitoring of localized damage patterns. However, this temporal advantage is accompanied by spectral limitations, as the absence of SWIR bands restricts the direct application of conventional burn indices and threshold-based approaches that rely on SWIR sensitivity to char and moisture loss \cite{giglio2016collection}.

This spatial–temporal–spectral trade-off motivates the adoption of an unsupervised variational autoencoder (VAE) framework. Rather than relying on predefined spectral indices or labeled training data, the VAE learns latent representations from the available RGB and Near-Infrared (NIR) bands, enabling detection of fire-related anomalies through changes in feature distributions between pre- and post-event imagery. In settings where timely updates are required and labeled datasets are limited, such an unsupervised approach leverages PlanetScope’s high revisit frequency while minimizing its spectral constraints. This methodological choice is therefore consistent with the objective of near–real-time conflict monitoring. By independently evaluating spatial partitions, the VAE framework has a linear time complexity with respect to the number of input tiles, making daily monitoring of massive regions like Darfur computationally feasible \cite{mateo2021towards, racek2025unsupervised}.

This work highlights the importance of interdisciplinary integration. Verification of detected incidents, uncertainty quantification, and operational relevance require input from peace and conflict research, humanitarian studies, and remote sensing expertise. Prior work emphasizes that automated satellite detection alone does not produce actionable insight; coordinated approaches are needed to interpret results, ensure reliability, and support evidence-based decision-making in conflict-affected contexts \cite{ghioni2024open, duursma2023peacekeeping}.

\section{Conclusion}
\label{sec5}

A civil war in Sudan, primarily between the Sudan Armed Forces (SAF) and the Rapid Support Forces (RSF), has caused widespread civilian casualties and displacement \cite{ahmed2025calling}, with over 1,400 violent incidents targeting civilians, resulting in 61,000 war-related deaths and 12 million displaced people \cite{rothbart2025sudan}. This study demonstrates that conflict-related fire damage in Sudan can be detected in near–real-time using PlanetScope imagery and latent representation–based change detection. The VAE-based approach demonstrated improved detection capabilities over traditional baselines, with 4-band imagery providing the most relevant information and additional bands or short temporal sequences yielding modest improvements. 

This study presents an approach integrating remote sensing, unsupervised learning, and conflict analysis to support humanitarian assessment in data-scarce environments. We acknowledge several limitations of this work. First, the analysis is based on a limited number of case studies within Sudan, which may constrain the generalizability of the findings across broader geographic and conflict contexts. Second, the comparative evaluation is limited to established pixel-level change detection methods grounded in spectral distance and transformation-based techniques, and the reliance on commercial Planet Labs PlanetScope imagery introduces challenges related to data accessibility, cost, and reproducibility. Furthermore, all quantitative results are contingent upon the availability of cloud-free optical imagery and accurate co-registration between pre- and post-incident scenes, while operational latency and system performance under cloud-covered conditions were not evaluated. Transient phenomena such as smoke plumes are inconsistently observed in optical imagery due to revisit timing and atmospheric variability, and thus are not a reliable signal for systematic detection. The proposed framework primarily relies on persistent surface changes (e.g., burn scars), which provide more stable indicators.

Despite these limitations, the findings indicate that lightweight, latent representation–based methods offer a practical approach for near–real-time monitoring of conflict-related fire damage, highlighting a trade-off between spectral richness, temporal availability, and feasibility in settings where higher-resolution or ground-based data are scarce and operationally impractical to acquire in real-time.

\section*{Acknowledgements}

The researchers involved in the SCO project include the authors of this article as well as the following analysts: Moneim Adam, Mathieu Bere, Hind Fadul, Anusha Srirenganathanmalarvizhi, Zifu Wang, David Wong and Chaowei Yang.

This project was supported by computing resources provided by the Office of Research Computing at George Mason University (URL: \url{https://orc.gmu.edu}) and funded in part by grants from the National Science Foundation (Awards Number 1625039,  2018631, and 2109647).

\section*{Funding}

The research conducted in this article was funded by the United States Department of State and contracted by The MITRE Corporation.

\section*{Conflicts of Interest}

The authors declare no conflicts of interest.

\section*{Data availability}

The data that support the findings of this study are available upon a reasonable request from the corresponding author. The data are not publicly available due to privacy or licensing restrictions.

\section*{Code availability}

The code is available in a GitHub repository at \url{https://github.com/heykuldip/nrt-conflict-monitoring}

\bibliographystyle{elsarticle-num-names}
\biboptions{sort&compress}

\end{document}